\documentclass{ecai} 

\usepackage{latexsym}
\usepackage{amssymb}
\usepackage{amsmath}
\usepackage{amsthm}
\usepackage{booktabs}
\usepackage{enumitem}
\usepackage{graphicx}
\usepackage{color}

\usepackage{booktabs}
\usepackage{multirow}
\usepackage{array}
\usepackage{makecell}
\usepackage{graphicx} 
\usepackage{longtable}
\usepackage{amsmath}
\usepackage{cleveref}

\usepackage{wrapfig}
\newcommand\ie{\textit{i.e.}}

\usepackage{amsfonts}
\usepackage{ulem}

\def\R{\mathbb{R}}
\def\tr{\mathrm{tr}}
\newcommand\il[1]{\langle #1 \rangle}
\def\our{LoRA-XS}
\usepackage{bbm}
\def\1{\mathbbm{1}}

\usepackage{amsthm}
\theoremstyle{remark}
\newtheorem{observation}{Observation}[section]

\DeclareMathOperator*{\argmin}{arg\,min}

\newtheorem{theorem}{Theorem}

\newcommand{\BibTeX}{B\kern-.05em{\sc i\kern-.025em b}\kern-.08em\TeX}

\begin{document}


\begin{frontmatter}


\paperid{8303} 



\title{LoRA-XS: Low-Rank Adaptation with Extremely Small Number of Parameters}


\author[A]{\fnms{Klaudia}~\snm{Bałazy}\footnote{Equal contribution. Correspondence to: klaudia.balazy@doctoral.uj.edu.pl\\Accepted at ECAI 2025}}
\author[B]{\fnms{Mohammadreza}~\snm{Banaei}\footnotemark}
\author[B]{\fnms{Karl}~\snm{Aberer}} 
\author[A]{\fnms{Jacek}~\snm{Tabor}} 

\address[A]{Jagiellonian University}
\address[B]{EPFL}

\begin{abstract}
The growth of large language models underscores the need for parameter-efficient fine-tuning. Despite its popularity, LoRA encounters storage and computational challenges when deploying multiple task- or user-specific modules. To address this, we introduce LoRA-XS, a novel fine-tuning method backed by a theoretical derivation. LoRA-XS drastically reduces trainable parameters by incorporating a small, trainable weight matrix between frozen low-rank matrices derived from the Singular Value Decomposition of pre-trained weights. This design enables LoRA-XS to reduce storage requirements by over 100x in 7B models compared to LoRA. Additionally, unlike other methods, LoRA-XS imposes no lower bound on trainable parameters – it can scale from a single parameter per module to arbitrarily large values, adapting to any storage or computational constraint. Evaluations on GLUE, GSM8K, MATH, and commonsense reasoning benchmarks across different model scales reveal that LoRA-XS consistently outperforms or matches LoRA and VeRA in accuracy, offering unmatched parameter efficiency. Our ablation studies highlight the significance of singular vectors in transformer weights, establishing LoRA-XS as a powerful, storage-efficient solution for scaling and personalizing large language models.
\end{abstract}


\end{frontmatter}







\section{Introduction}
\label{introduction}

In recent years, the development of large language models (LLM) has revolutionized the field of natural language processing (NLP), enabling unprecedented performance across various tasks. However, these state-of-the-art models often come with a huge number of parameters, presenting significant challenges for fine-tuning and adaptation to specific downstream tasks. Modifying and storing these immense models introduces computational and storage challenges. 

To address these challenges, Parameter-Efficient Fine-Tuning (PEFT) methods have emerged as a promising solution by fine-tuning only a small subset of parameters~\citep{houlsby2019parameter,lora,lester2021power,li2021prefix,zaken2021bitfit}. Among them, LoRA (Low-Rank Adaptation)~\citep{lora} has gained popularity for its strong generalization and no impact on inference latency. However, even LoRA can require considerable storage and computational resources when deployed at scale. For instance, if we would consider a scenario where personalized adapters are needed for multiple users (such as in personalized educational models), adapting GPT-3~\cite{brown2020language} with LoRA (rank 16, query, and value matrices) requires 144MB per checkpoint, resulting in 144TB of storage for 1 million personalized models.

Following LoRA, many successors have been proposed to further reduce the number of parameters and improve efficiency~\citep{kopiczko2023vera,dora,adalora}. One such recent method is VeRA~\citep{kopiczko2023vera}, which reduces the number of trainable parameters by using a single pair of low-rank frozen matrices shared across all layers while learning small scaling vectors instead. Although VeRA improves parameter efficiency, its parameter count remains dependent on the model hidden dimensions, which becomes increasingly significant for larger language models.\footnote{For example, while early transformer models like BERT have a hidden dimension of 768, the recent GPT-3 model has a hidden dimension of 12288, which directly affects the trainable parameter count for LoRA and VeRA methods.} This dependency can result in substantial storage and computational requirements as model sizes continue to grow. 

In this paper, we propose LoRA-XS~(\textbf{Lo}w-\textbf{R}ank \textbf{A}daptation with e\textbf{X}tremely \textbf{S}mall number of parameters), a highly parameter-efficient fine-tuning method built on theoretical insights\footnote{Code: https://github.com/MohammadrezaBanaei/LoRA-XS}. LoRA-XS achieves superior performance to LoRA and VeRA while drastically reducing trainable parameters. Notably, the parameter count in LoRA-XS is independent of the model's hidden dimensions, offering a transformative approach to parameter-efficient fine-tuning. Given the previous example, adapting GPT-3 with LoRA-XS (rank 16) for 1 million personalized models requires just 96GB of storage, compared to LoRA's 144TB (a 1500x reduction).

\begin{figure}[ht]
  \centering
    \includegraphics[width=1.0\columnwidth]{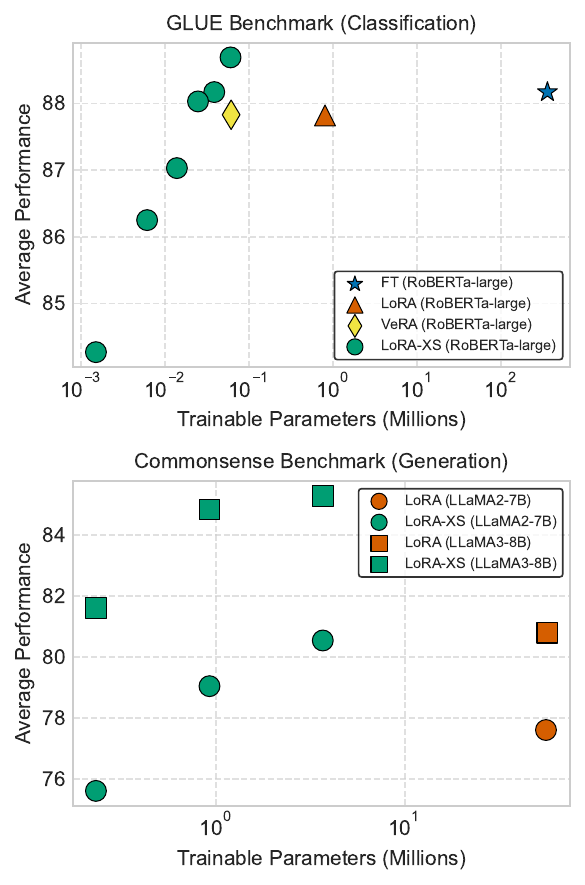}
    \caption{Average performance for a given number of trainable parameters across different adaptation methods and model scales. The top plot shows RoBERTa-large performance (million-scale LLM) on a subset of GLUE tasks, while the bottom plot presents the performance on commonsense reasoning tasks for LLaMA2-7B and LLaMA3-8B (billion-scale LLMs). LoRA-XS points correspond to various ranks (please refer to~\Cref{tab:gluebenchmark} and~\Cref{tab:commonsense} for the details). LoRA-XS consistently outperforms other methods in both parameter efficiency and average performance. Unlike other approaches, LoRA-XS provides greater flexibility in reducing the number of trainable parameters, as it is not constrained by model dimension, enabling more efficient adaptation without a lower bound. These results establish LoRA-XS as a scalable solution for parameter-efficient fine-tuning.}
    \label{fig:generic_perf_parameter}
\end{figure}

Moreover, with LoRA-XS, we can precisely control the number of additional parameters, allowing for flexible memory usage (see \Cref{fig:generic_perf_parameter}). This flexibility is particularly beneficial for increasingly larger models, where traditional methods impose a certain minimum number of additional parameters. Furthermore, LoRA-XS retains the core advantages of LoRA, such as not requiring any modifications to the model architecture and introducing no additional latency during inference, making it an efficient solution for practical deployment.

LoRA-XS achieves the extreme parameter efficiency by setting LoRA's projection matrices using Singular Value Decomposition (SVD) of the pre-trained module weights and keeping them frozen during training (see Figure~\ref{fig:ourlora}). The only trainable parameter of LoRA-XS is an $r \times r$ matrix (\ie, $R$) between the frozen LoRA projection matrices, where $r$ denotes LoRA's rank.

Fixing these matrices during training transforms LoRA-XS into a \textit{latent editing} approach, with the matrix $R$ requiring only $r^2$ parameters. LoRA-XS consistently delivers superior performance compared to LoRA and recent methods like VeRA across various benchmarks and model scales. We evaluate LoRA-XS on GLUE~\citep{wang2018glue} for natural language understanding, GSM8K~\citep{gsm8k}, and MATH~\citep{mathdataset} for mathematical reasoning, as well as eight commonsense reasoning datasets (see~\Cref{experiments}).

We also conduct an extensive ablation study, revealing the essential role of singular vectors in transformer weights, which highlights the core mechanism behind LoRA-XS's efficiency (see~\Cref{sec:mainablation}). Our findings demonstrate that LoRA-XS is both remarkably parameter-efficient and a transformative solution for scaling and tailoring large language models to unprecedented levels. Moreover, LoRA-XS method is orthogonal to other efficiency techniques, such as pruning, quantization, or dynamic rank adaptation, and can complement them to achieve additional memory savings.

In summary, our contributions are as follows:
\begin{itemize}
    \item We introduce LoRA-XS, a novel fine-tuning method grounded in theoretical derivation, achieving over 100x reduction in trainable parameters for large-scale models without sacrificing performance.

    \item LoRA-XS consistently outperforms LoRA and recent methods like VeRA across various model sizes and diverse tasks, including GLUE, GSM8k, MATH, and eight commonsense reasoning benchmarks.  

    \item Unlike existing LoRA variants, LoRA-XS enables precise control over memory, allowing scaling trainable parameters from as small as one parameter per module to arbitrarily large number of parameters (see~\Cref{fig:generic_perf_parameter}).

\end{itemize}

\section{Related work}
\label{related_work}

\paragraph{Efficient Adaptation} Recently, many adapter-based fine-tuning methods have been proposed, introducing adapter modules into transformer models~\citep{vaswani2017attention}. These modules can either be introduced as extra \textit{adapter} layers into the transformer block~\citep{houlsby2019parameter,liu2022few,pfeiffer2020adapterfusion}, or as an additional set of parameters modifying input layer activations \citep{asai2022attempt,lester2021power,li2021prefix,liu2023gpt}. While these approaches add relatively few trainable parameters, they increase latency during online inference, posing challenges for large-scale production deployments.

\paragraph{Low-rank Adaptation} Low-rank adaptation of transformer models, proposed by LoRA~\citep{lora}, offers a strong alternative to previous PEFT methods, where the generalization performance is competitive to full fine-tuning while not introducing any further latency during inference phase. Building upon the LoRA method, there has been many recent efforts to improve its learning curve~\citep{dora,hayou2024lora+,meng2024pissa}, reduce the trainable parameters~\citep{adalora,kopiczko2023vera,renduchintala2023tied}, or even training it with quantized pre-trained weights to improve memory footprint during training~\citep{dettmers2024qlora,li2023loftq}. Our method, LoRA-XS, falls in the second category, where we aim to significantly reduce trainable parameters while performing competitively to LoRA over various benchmarks and different model scales. 

\paragraph{Parameter-constrained LoRA variants} Several recent works propose variants of LoRA that reduce the number of trainable parameters while maintaining competitive performance. AdaLoRA~\citep{adalora} investigates a dynamic rank adjustment for different modules' low-rank matrices as opposed to uniform parameter allocation in LoRA. Tied-LoRA~\citep{renduchintala2023tied} improves the parameter efficiency by tying LoRA matrices across all layers of the transformer model. VeRA~\citep{kopiczko2023vera}, which is closely related to our work, shares randomly initialized frozen LoRA matrices across layers and adds trainable scaling vectors. However, unlike VeRA, LoRA-XS initializes its low-rank matrices using the SVD of the pre-trained model weights, providing both theoretical justification (see~\Cref{theory}) and strong empirical performance (see~\Cref{sec:mainablation}). Additionally, LoRA-XS’s number of trainable parameters is independent of the model's hidden dimensions, allowing for a significant reduction in parameters, particularly in large-scale models.

\section{Method}
\label{method}

\begin{figure*}[t]
  \centering
  \includegraphics[width=0.8\textwidth]{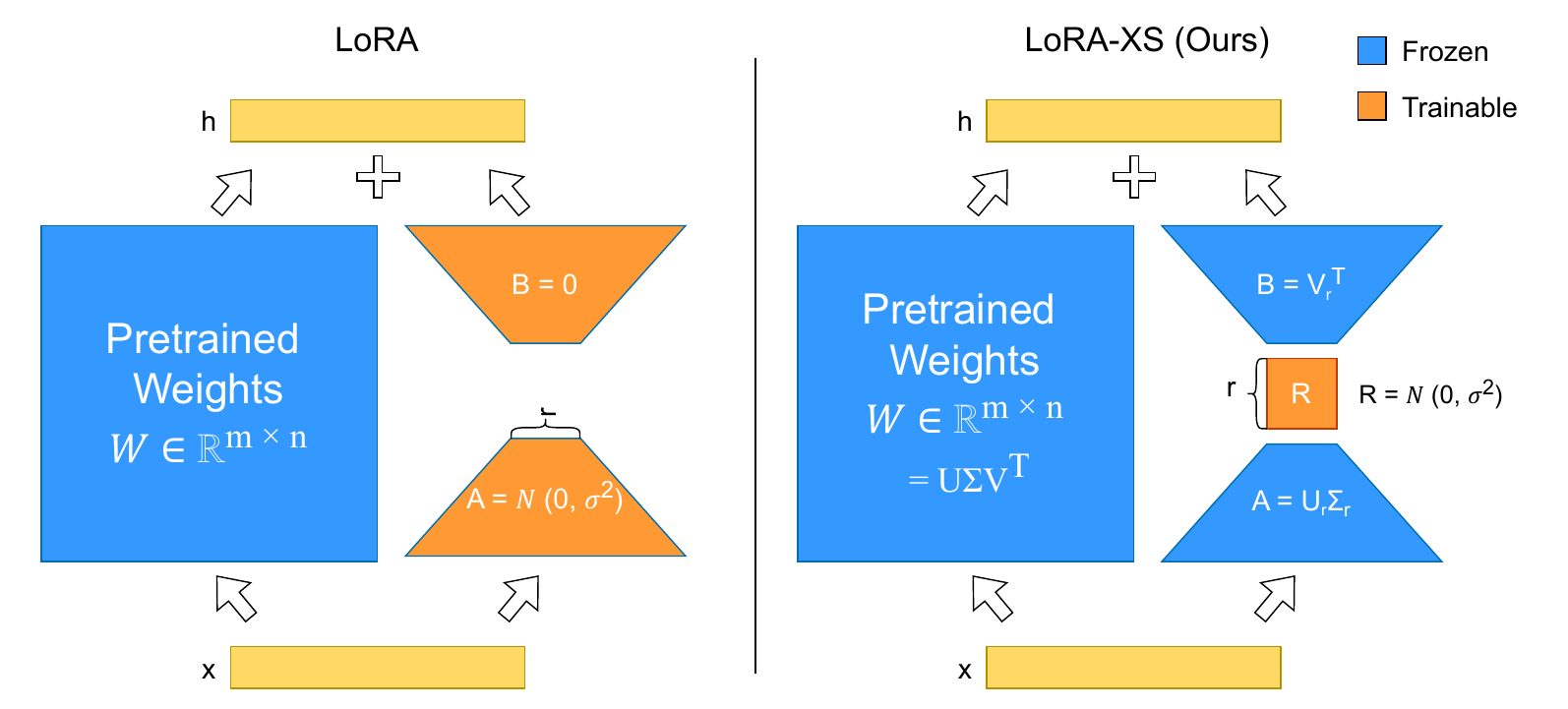}

    \caption{A visual comparison of the LoRA and LoRA-XS techniques. The key innovation of LoRA-XS is its small trainable matrix $R$, positioned between two frozen low-rank matrices derived from the truncated SVD of pre-trained weights. Unlike other methods, where adaptation scales with model dimensions, LoRA-XS enables flexible memory allocation, allowing trainable parameters to range from a single parameter per module to arbitrarily large values.}

    \label{fig:ourlora}
\end{figure*}

This section introduces LoRA-XS (\textbf{Lo}w-\textbf{R}ank \textbf{A}daptation with e\textbf{X}tremely \textbf{S}mall number of parameters), a novel method designed to improve the parameter efficiency of fine-tuning large language models by leveraging insights from low-rank adaptation. Building on LoRA's core ideas, our approach addresses scalability and storage issues while maintaining highly competitive performance.

In recent years, LoRA~\citep{lora} has been pivotal in parameter-efficient tuning by introducing low-rank matrices for adaptation, significantly lowering the number of trainable parameters. However, deploying LoRA at scale, particularly across large-scale task-specific or user-specific models, can substantially increase the required storage. As modern models grow in size and complexity, the need for even more parameter-efficient tuning strategies becomes a crucial issue.

LoRA demonstrated that its low-rank weight updates (\ie, $\Delta W$) align with certain \textit{directions} already present in the model's weights. Building on this insight and our theoretical derivation (see~\Cref{theory}), we propose initializing the LoRA adaptation matrices ($A$ and $B$ in \Cref{fig:ourlora}) using the top singular vectors from the SVD of the pre-trained weight matrix $W$. These matrices are kept fixed during training. To introduce flexibility, we add a trainable $r \times r$ matrix $R$ between $A$ and $B$, making $R$ the only learnable component. This drastically reduces the number of trainable parameters, while keeping the parameter count independent of the model's dimensions. \Cref{fig:ourlora} provides an overview of the LoRA-XS method, highlighting its differences from the original LoRA framework.

\subsection{Theoretical foundations of LoRA-XS}
\label{theory}

We provide a detailed theoretical derivation of \our{} in \Cref{appendix:theory}, establishing its mathematical foundations and motivating our choice of adaptation subspaces. For readers primarily interested in practical aspects, we summarize the main insights below.

A key question in parameter-efficient fine-tuning is: \textit{What is the optimal low-dimensional subspace for adapting model weights?} Ideally, updates should be restricted to the most informative directions, maximizing expressivity while minimizing trainable parameters.

In typical low-dimensional adaptation of pre-trained weights in deep models, we need to determine the form of the subspace in which the adapted weights should reside. For example, in the well-known LoRA approach, the weight matrix $W$ is modified by adding the product of two low-rank matrices:
\begin{equation}
    W' = W + AB,
\end{equation}
where $A \in \mathbb{R}^{n \times r}$ and $B \in \mathbb{R}^{r \times n}$ are trainable matrices of rank $r$, with $A$ initialized from a random Gaussian distribution and $B$ initialized to zero.

In our approach, we theoretically show that, in a large class of problems, the optimal modification of weights is given by:
\begin{equation}
    W' = W + U \Sigma R V^T,
\end{equation}
where $U, \Sigma, V$ are obtained from the SVD of $W$, and $R \in \mathbb{R}^{r \times r}$ is a trainable matrix. This formulation ensures that parameter updates align with the dominant singular directions of \(W\), thereby enhancing adaptation efficiency. Consequently, \our{} significantly outperforms other methods (see~\Cref{experiments}). Furthermore, our method enables efficient orthogonal projection onto the adaptation subspace via a closed-form expression, facilitating post-training dimensionality reduction (see~\Cref{sec:mainablation}).

Our experiments empirically validate this theoretical intuition, demonstrating that initializing \our{} with SVD improves fine-tuning performance across multiple tasks (see~\Cref{experiments} and~\Cref{sec:mainablation}). For a complete derivation, refer to \Cref{appendix:theory}.

\subsection{Theoretical derivation of LoRA-XS}
\label{appendix:theory}

In this section, we derive the theoretical foundations of our approach.

We begin by considering a neural network with transformer architecture. Let $W \in \mathbb{R}^{n \times n}$ be a square weight matrix for an arbitrary linear layer in this network. Our goal is to adapt the weights to new tasks by applying a correction $\Delta W$, which we want to constrain to a lower-dimensional subspace of $\mathbb{R}^{n \times n}$. We aim to show how to choose such a subspace to allow for significant flexibility and to work effectively for future gradient adaptations.

In the standard LoRA method, the subspace used for adaptation is characterized as the set of all matrices of the form $AB$, where $A \in \mathbb{R}^{r \times n}$ and $B \in \mathbb{R}^{n \times r}$, resulting in a space with dimension equal to $2nr$. However, in our proposed \our{} framework, we introduce a more general parametrization of subspaces, which allows for a dimension $r^2$, where $r$ can range from 1 to $n$. Compared to LoRA, which has at least dimension of $2n$, our approach can use an arbitrarily small amount of memory (see~\Cref{fig:generic_perf_parameter}).

Formally, given fixed orthogonal matrices $A \in \mathbb{R}^{r \times n}$ and $B \in \mathbb{R}^{n \times r}$, we define the subspace $S^r_{A,B}$ as:
$$
S^r_{A,B}=\{AXB^T:X \in \R^{r \times r}\}.
$$
This subspace has dimension $r^2$ and allows for a flexible choice of $r$ to adjust the dimensionality. Moreover, we can easily compute the orthogonal projection onto $S^r_{A,B}$. Namely,
$$
p_{A,B}(X)=A[A^TXB]B^T \text{ for }X \in \R^{n \times n},
$$
is the orthogonal projection with respect to Frobenius scalar product in the space of matrices on $S^r_{A,B}$
(the proof of this result is presented in~\Cref{appendix:theory_proof}). This projection can be useful when we want to project a full gradient onto the space $S^r_{A,B}$.

\paragraph{Main idea behind \our{}}
The problem that \our{} aims to solve is how to choose the matrices $A$ and $B$ such that the following optimization procedures yield similar weights:
\begin{itemize}
    \item fine-tuning the network's weights without any restrictions,
    \item fine-tuning the network's weights restricted to $S^r_{A,B}$.
\end{itemize}
We demonstrate that, under reasonable assumptions, the optimal matrices $A$ and $B$ are obtained through truncated SVD on the initial weights $W$.

Consider the fine-tuning of the model. Assume that we have a pre-trained weight matrix $W \in \R^{n \times n}$, and that we want to find the space $S^r_{A,B}$ in which the modification would be optimal for the further fine-tuning. Let $G_1,\ldots,G_k$ denote the gradients computed for the mini-batches. During SGD optimization with learning rate $h$, we would arrive at the weights:
$$
W+\Delta W=W+hG_1+\ldots+hG_k.
$$

If we have chosen our subspace $S^r_{A,B}$ such that $\Delta W$ is close to it, we can transition from $S^r_{A,B}$ to $\Delta W$ during fine-tuning. More precisely, if $\Delta W \in S^r_{A,B}$, then $p_{A,B}(\Delta W)=\Delta W$, and by applying the orthogonal projection in our model, we get:
\begin{multline}
W+h p_{A,B}(G_1)+\ldots+hp_{A,B}(G_k)= \\ 
W+p_{A,B}(hG_1+\ldots+hG_k)= \\
W+p_{A,B} (\Delta W)=W+\Delta W.
\end{multline}

\begin{theorem}
Let $G$ denote the mean gradient:
$G=\frac{1}{k}\sum_{i=1}^k G_i$.
Let us apply truncated SVD decomposition on $G$ to obtain $U_r,\Sigma_r,V_r$. Then
$$
U_r,V_r=\argmin_{A,B} d(G;S^r_{A,B}),
$$ 
where $d$ denotes the distance in the Frobenius norm.
\end{theorem}

\begin{proof}
Observe that every element of $S^r_{A,B}$ is trivially of rank at most $r$. By the matrix approximation lemma, known as the Eckart–Young–Mirsky theorem, we know that the optimal approximation in Frobenius norm of matrix $G$ in the rank $r$ matrices is given by $U_r,\Sigma_r,V_r^T$, where $G=U\Sigma V^T$ is the SVD decomposition of $G$.  
\end{proof}

To extend this result to \our{}, we make the additional assumption that the gradients during fine-tuning do not essentially diverge from those observed during pre-training. In practice, this is often the case, as fine-tuning tasks are typically similar to pre-training tasks, resulting in a shift in the distribution of gradients rather than an entirely new distribution. Indeed, the efficiency of \our{}, as demonstrated in our experiments, supports this assumption (see~\Cref{experiments}). Moreover, our ablation study on \our{} initialization shows a significant benefit when using the SVD of the weights for tasks aligned with language modeling. For SST-2, this initialization provided similar accuracy to random initialization (see~\Cref{main:ablation:svdinit} and \Cref{appendix:ablation_loraxs_init}), which may stem from the fact that this task is not well aligned with language modeling compared to the other tested tasks, such as CoLA, MRPC, and QNLI.

\begin{observation}
Consider a deep neural network $\Phi$ pre-trained on a dataset $x_1,\ldots,x_k$, with weights $W$. Let $G_i=\nabla \Phi(x_i)$ be the gradients of the pre-trained model. Then,
$$
W \sim \frac{1}{k}\sum_{i=1}^k G_i.
$$
This follows from the fact that at the final phase of the neural networks training, its weights stabilize. Consequently, at the final phase of training, the previous gradients $\tilde G_t$ computed at iteration $t$ are close to the ones computed for the fully trained model. Then for $W_t$ denoting the weights of the model in the step $t$, we get:
\begin{multline}
W_T=W_t+h \tilde G_t+\ldots+h\tilde G_{T-1}
\approx \\
W_t+h G_t+\ldots+ h G_{T-1}.
\end{multline}

Consequently, the gradients accumulate and represent the largest part of the sum:
$$
W_T \approx h (G_t+\ldots+G_{T-1}). 
$$
Since gradients come from the same distribution we obtain $W=W_T \sim \frac{1}{k}\sum_i G_i$.
\end{observation}

This leads us to the formulation of \our{}.
Suppose we aim to obtain a good $d$-dimensional approximation for the optimal gradient training of a linear layer with weights $W$ in a pre-trained transformer model, where $d=r^2$. To do this, we apply truncated (to $r$ eigenvalues) SVD to weights $W$ obtaining $U_r,\Sigma_r,V_r^T$. Then, to make an update with learning rate $h>0$:
\begin{itemize}
    \item compute the gradient $G_i$ for the model over the consecutive $i$-th mini-batch,
    \item update the weights by taking the projection of the computed gradient on the space $S^r_{U_r,V_r}$:
    $$
    \Delta W_i=h \cdot U_r[U_r^TG_iV_r]V_r^T.
    $$
\end{itemize}
In practice, the projection is not necessary, as we can work directly in the space $S^r_{U,V}$. Thus, we can compute the gradient with respect to $r \times r$ dimensional update space:
$$
W+U_rRV_r^T \text{ for trainable }R \in \R^{r \times r}.
$$
Observe that, due to the truncated SVD decomposition, the matrix $R$ has a size of $r \times r$. The next subsection provides a more direct description of our method. Additionally, please refer to the experiments and ablation sections (see~\Cref{experiments} and~\Cref{sec:mainablation}), which empirically support our theory of \our{}. One additional factor we observed empirically is that it is beneficial to rescale $R$ by the components of $\Sigma_r$ (see~\Cref{sec:mainablation} and~\Cref{ablation:sigmaornotinit}). Thus, in \our{}, we optimize in the space $W + U_r\Sigma_r R V_r^T$, where $R$ is the trainable matrix with $r \times r$ coefficients.

\subsection{Formulation of LoRA-XS}

The main idea behind the LoRA-XS is to modify the adaptation process by introducing a small square matrix $R \in \mathbb{R}^{r \times r}$ between frozen LoRA matrices that are set using truncated SVD of the pre-trained weight matrix $W \in \mathbb{R}^{m \times n}$.\footnote{From now on, to maintain consistency with common notation conventions and our LoRA-XS code, we will work in the transposed space, where vectors are represented as rows, and the multiplication of a vector $x$ by a matrix $A$ is expressed as $xA$. Consequently, $W$ will formally denote the transposed weight matrix.
}

The traditional LoRA forward path for an input $x \in \mathbb{R}^{n}$ can be formulated as:
$$
h = x W +  x ~\Delta W = x W + x A B, 
$$
where $\Delta W \in \mathbb{R}^{m \times n}$ is the low-rank weight update.
The matrices $A \in \mathbb{R}^{m \times r}$ and $B \in \mathbb{R}^{r \times n}$ are low-rank matrices with $r \ll \min(m, n)$. During training, $W$ is kept frozen, and $A$ and $B$ are the trainable parameters.

In LoRA-XS, we improve parameter efficiency by introducing a small trainable matrix $R \in \mathbb{R}^{r \times r}$, while keeping matrices $A$ and $B$ frozen, modifying the forward path to:
$$
h = x W + x~\Delta W  = x W +  x A R B. 
$$
Here, $A$ and $B$ are set using the truncated SVD of the original weight matrix $W$. Formally, the SVD of $W$ is given by:
$$
W = U \Sigma V^T,
$$

where $U \in \mathbb{R}^{m \times m}$, $\Sigma \in \mathbb{R}^{m \times n}$, and $V \in \mathbb{R}^{n \times n}$. We set (frozen) matrices $A$ and $B$ as:
$$
A = U_r \Sigma_r \quad \text{and} \quad B = V_r^T,
$$
where $U_r \in \mathbb{R}^{m \times r}$ and $V_r \in \mathbb{R}^{n \times r}$ contain the left/right singular vectors corresponding to the top $r$ singular values, and $\Sigma_r \in \mathbb{R}^{r \times r}$ is a diagonal matrix that contain top $r$ singular values of $\Sigma$.

Our newly introduced $R$ matrix is initialized with a Gaussian distribution $N(0, \sigma^2)$, where $\sigma$ is set to a small but non-zero value\footnote{In all LoRA-XS experiments, we set $\sigma$ to $10^{-5}$.}. This ensures that we start fine-tuning with a model almost identical to the original pre-trained model. During fine-tuning, the matrices $A$ and $B$ are kept frozen, and only $R$ is updated, significantly reducing the number of trainable parameters.

Compared to other LoRA variants, LoRA-XS provides better control over the number of trainable parameters, allowing for more flexible storage requirements for the fine-tuned models. This flexibility is particularly beneficial for increasingly larger models, where traditional methods are often limited by model's hidden dimensions. Similar to LoRA and its successors, LoRA-XS does not introduce any extra computational overhead or latency during inference, as this module can be merged into the original matrix post-training.

We make the following observation on the parameter efficiency of LoRA-XS compared to LoRA and VeRA methods (please refer to~\Cref{appendix:param_eff_loraxs} for the details).

\textbf{Observation:} LoRA-XS demonstrates superior parameter efficiency over LoRA and VeRA.

Let's consider a transformer model with $L$ fine-tuned layers, each consisting of $q$ number of $W \in \mathbb{R}^{n \times n}$ matrices. As the model dimension $n$ becomes very large compared to the rank $r$, the benefit of LoRA-XS over LoRA and VeRA becomes more pronounced. Specifically for large $n$:
$$
    \frac{P_{\text{LoRA}}}{P_{\text{LoRA-XS}}} \approx \frac{2n}{r}
    \quad \text{and} \quad
    \frac{P_{\text{VeRA}}}{P_{\text{LoRA-XS}}} \approx \frac{n}{r^2}.
$$
This indicates that for large models, LoRA and VeRA require significantly more parameters than LoRA-XS, with the difference growing linearly with $n$ (\ie, model's hidden dimension). This makes LoRA-XS especially suitable for fine-tuning models where parameter efficiency is crucial.

\section{Experiments}
\label{experiments}

\begin{table*}[ht!]
    
    \caption{
    RoBERTa-large performance on six GLUE tasks with different adaptation methods. We report Matthews correlation for CoLA, Pearson correlation for STS-B, and accuracy for the remaining tasks. Full fine-tuning, LoRA, and VeRA results are taken from prior works~\citep{lora,kopiczko2023vera}. LoRA-XS matches or surpasses baselines while offering significantly better parameter efficiency. We bold the best scores among parameter-efficient methods. Since in this experiment our primary goal is to compare LoRA-XS with VeRA, we underscore average scores where LoRA-XS outperforms VeRA while maintaining greater parameter efficiency.}
    \centering
    \vskip 0.1in
    \resizebox{\textwidth}{!}{
    \begin{tabular}{@{}c|cc|cccccc|c@{}}
         
         \midrule
         Method & \makecell{\# Trainable \\ Parameters} & Rank & SST-2 & MRPC & CoLA & QNLI & RTE & STS-B & Avg. \\
         \hline

         FT & 355,000K & - & 96.4 & 90.9 & 68.0 & 94.7 & 86.6 & 92.4 & 88.17 \\
         \hline

         LoRA & 800K & 8 &  96.2 $\pm$ 0.5 & 90.2 $\pm$ 1.0 & 68.2 $\pm$ 1.9 & \textbf{94.8} $\pm$ 0.3 & 85.2 $\pm$ 1.1 & \textbf{92.3} $\pm$ 0.5 & 87.82 \\

         VeRA & 61K & 256 & 96.1 $\pm$ 0.1 & 90.9 $\pm$ 0.7 & 68.0 $\pm$ 0.8 & 94.4 $\pm$ 0.2 & 85.9 $\pm$ 0.7 & 91.7 $\pm$ 0.8 & 87.83 \\
        \hline
        \multirow{6}{*}{LoRA-XS} & \uline{\textbf{60K}} & 25 & \textbf{96.3} $\pm$ 0.2 & \textbf{91.2} $\pm$ 0.8 & \textbf{68.6} $\pm$ 0.8 & 94.3 $\pm$ 0.2 & \textbf{89.5} $\pm$ 0.5 & 92.2 $\pm$ 0.1 & \uline{\textbf{88.69}} \\

         & \uline{38.4K} & 20 & 95.9 $\pm$ 0.3 & 90.4 $\pm$ 0.4 & 68.1 $\pm$ 1.2 & 94.1 $\pm$ 0.2 & 88.8 $\pm$ 0.2 & 91.8 $\pm$ 0.2 & \uline{88.17} \\

         & \uline{24.6K} & 16 & 95.9 $\pm$ 0.2 & 90.7 $\pm$ 0.4 & 67.0 $\pm$ 1.2 & 93.9 $\pm$ 0.1 & 88.8 $\pm$ 0.3 & 92.0 $\pm$ 0.1 & \uline{88.03} \\
        
         & 13.8K & 12 & 95.9 $\pm$ 0.3 & 90.2 $\pm$ 0.3 & 65.5 $\pm$ 0.9 & 93.3 $\pm$ 0.5 & 87.7 $\pm$ 0.7 & 91.4 $\pm$ 0.1 & 87.03 \\

         & 6.1K & 8 & 95.3 $\pm$ 0.3 & 88.5 $\pm$ 0.6 & 64.4 $\pm$ 0.8 & 92.5 $\pm$ 0.1 & 86.3 $\pm$ 0.6 & 90.6 $\pm$ 0.3 & 86.25 \\

         & 1.5K & 4 & 94.8 $\pm$ 0.3 & 87.8 $\pm$ 0.3 & 60.5 $\pm$ 1.5 & 90.9 $\pm$ 0.3 & 82.7 $\pm$ 0.5 & 88.9 $\pm$ 0.2 & 84.27 \\
         
         \bottomrule
    \end{tabular}
    }
    \label{tab:gluebenchmark}
\end{table*}

\begin{table*}[ht!]
     \caption{Accuracy evaluation of fine-tuned LLaMA2-7B and LLaMA3-8B models across eight commonsense reasoning datasets. The rank for LoRA is 32, and the ranks for LoRA-XS are 128, 64, and 32, respectively. LoRA-XS consistently outperforms LoRA baseline while using only a small fraction of its trainable parameters, demonstrating exceptional parameter efficiency. Notably, even the smallest LoRA-XS model (with 0.23M trainable parameters) can outperform LoRA on some tasks. These results highlight LoRA-XS's ability to generalize effectively to large-scale models while achieving superior accuracy.}
    \centering
    \vskip 0.1in
    \resizebox{\textwidth}{!}{

\begin{tabular}{c|cccccccccc|c}
\toprule
Model & Method & \makecell{\# Trainable \\ Parameters} & BoolQ & PIQA & SIQA & \makecell{Hella-\\Swag} & \makecell{Wino-\\Grande} & ARC-e & ARC-c & OBQA & Avg. \\
\hline

\multirow{4}{*}{\makecell{LLaMA2 \\ 7B}}  & LoRA & 56M & 69.8 & 79.9 & 79.5 & 83.6 & 82.6 & 79.8 & 64.7 & 81.0 & 77.6 \\ 
\cline{2-12}
                            & \textbf{LoRA-XS} & \textbf{3.67M} & \textbf{72.1} & \textbf{83.7} & \textbf{80.5} & \textbf{86.0} & \textbf{83.9} & \textbf{86.0} & \textbf{71.2} & \textbf{81.0} & \textbf{80.5}  \\
                            & LoRA-XS & 0.92M & 70.4 & 83.0 & 80.2 & 82.3 & 82.7 & 84.4 & 68.6 & 80.8 & 79.0 \\
                            & LoRA-XS & 0.23M &  67.2 & 81.8 & 78.1 & 75.4 & 80.8 & 81.2 & 65.9 & 74.6 & 75.6  \\
\hline
\multirow{4}{*}{\makecell{LLaMA3 \\ 8B}}  & LoRA & 57M & 70.8 & 85.2 & 79.9 & 91.7 & 84.3 & 84.2 & 71.2 & 79.0 & 80.8  \\ 
\cline{2-12}
                            & \textbf{LoRA-XS} & \textbf{3.67M} & \textbf{73.2} & \textbf{88.3} & \textbf{82.5} & \textbf{94.1} & \textbf{87.1} & \textbf{90.9} & \textbf{80.5} & \textbf{85.6} & \textbf{85.3} \\
                            & LoRA-XS & 0.92M & 72.8 &	87.4 &	81.5 & 94.2 &	87.4 & 91.3 &	79.2 & 85.0 & 84.8 \\
                            & LoRA-XS & 0.23M & 66.6 & 85.8 & 79.4 & 90.1 & 85.2 & 87.0 & 76.5 & 81.8 & 81.6 \\

\bottomrule 
\end{tabular}
}

    \label{tab:commonsense}
\end{table*}

This section describes experiments evaluating the effectiveness of LoRA-XS. We begin by detailing the experimental setup and then then provide results for the GLUE benchmark~\citep{wang2018glue}, where we compare LoRA-XS with full fine-tuning, LoRA, and VeRA across six tasks. We explore various ranks for LoRA-XS to highlight their effect on performance and parameter efficiency.

We also report results on instruction tuning tasks with decoder-only language models. These experiments test LoRA-XS's ability to enable large language models to follow instructions with minimal parameter overhead. The first set of experiments involves training models on the MetaMathQA dataset~\citep{yu2023metamath} and evaluating them on the GSM8K~\citep{gsm8k} and MATH~\citep{mathdataset} benchmarks, focusing on mathematical reasoning. The second set evaluates commonsense reasoning in the same setting as \citet{hu2023llm}, using eight benchmarks to assess model performance.

\subsection{Experimental Setup}

For the GLUE benchmark experiments, we use the RoBERTa-large model~\citep{liu2019roberta} and explore different ranks for LoRA-XS, ranging from $r=4$ to $r=25$. This range allows us to examine the impact of varying numbers of trainable parameters on performance. For GLUE experiments, we add LoRA-XS modules to the Query, Value, Attention Output, and the Output Fully Connected weight matrices in all layers of the RoBERTa-large model. Hyperparameters were selected through grid search, and the chosen values are summarized in \Cref{tab:gluehyperparams}. Similar to previous efforts, when fine-tuning the model on MRPC, RTE and STS-B tasks, the model is initialized using weights fine-tuned on the MNLI task\footnote{The model is trained for 10 epochs on the MNLI dataset and then further fine-tuned on these three tasks.}~\citep{lora}.

For the commonsense reasoning experiments, we use LLaMA2 7B~\citep{touvron2023llama} and LLaMA3 8B~\citep{dubey2024llama} decoder-only models, fine-tune them on a mixture of eight sub-tasks\footnote{The training dataset is a collection of 170K commonsense reasoning samples derived from \citet{hu2023llm}.}, and then separately evaluate the fine-tuned models on the validation set of these eight datasets~(see Appendix~\ref{appendix:setup_instruction} for more details). Our training/evaluation setting follows prior work~\citep{hu2023llm} in order to have a fair comparison with LoRA as the baseline method.

For the mathematical reasoning experiments, we use the Mistral-7B-v0.1~\citep{jiang2023mistral} and Gemma-7B~\citep{team2024gemma} decoder-only models, training them on 100k samples of the MetaMathQA~\citep{yu2023metamath} dataset and evaluating on the GSM8K~\citep{gsm8k} and MATH~\citep{mathdataset} datasets.

For all instruction tuning experiments, LoRA-XS modules are added to the Query, Key, Value, Attention Output, and all three fully connected weight matrices. Each LoRA-XS module in our main experiments is initialized with the truncated SVD of the corresponding pre-trained weights $W$. Further details of the experimental setup are provided in the~\Cref{appendix:setup}.

\subsection{GLUE Benchmark}
\label{glue}

In~\Cref{tab:gluebenchmark}, we present the performance of the RoBERTa-large model evaluated on the GLUE benchmark leveraging full fine-tuning (FT) and parameter-efficient fine-tuning methods: LoRA, VeRA and LoRA-XS.

As shown in~\Cref{tab:gluebenchmark}, LoRA-XS with ranks of 25, 20, and 16 (corresponding to 60K, 38.4K, and 24.6K trainable parameters, respectively) outperforms baseline PEFT methods (LoRA and VeRA), achieving the highest average performance across the tested GLUE tasks. Notably, with a rank of 16, LoRA-XS achieves better accuracy than VeRA while having 2.5x less trainable parameters.

Following the baselines evaluation method, we conduct 5 runs with different seeds, recording the best epoch’s outcome for each run, and report the median of these results.
Similar to LoRA~\citep{lora} and VeRA~\citep{kopiczko2023vera}, we only include the added LoRA-XS modules in the calculation of trainable parameters count, excluding the classifier parameters for a clear comparison. 

We observe competitive performance even at extremely low ranks, highlighting LoRA-XS's parameter efficiency. Moreover, unlike LoRA and LoRA-XS, VeRA requires a high rank (256 in this experiment) to stay competitive, significantly increasing training FLOPs and GPU memory usage due to its large A and B matrices.

Interestingly, despite using only around 1,500 parameters at rank 4, LoRA-XS retains strong performance, with an accuracy drop of just about 4 percentage points. This highlights its ability to achieve competitive results while significantly reducing the number of trainable parameters.

\subsection{Instruction tuning}
\label{instruction_tuning}

\begin{table}[ht!]
  \caption{Instruction tuning performance on the GSM8K and MATH benchmarks for Mistral-7B and Gemma-7B. The rank for LoRA is 64, and the ranks for LoRA-XS are 128, 64, 32 and 16, respectively. LoRA-XS achieves superior or competitive results compared to LoRA and full fine-tuning (FT). For instance, on Mistral-7B, LoRA-XS with just 3.67M trainable parameters outperforms LoRA with 168M parameters, highlighting its ability to drastically reduce parameter count without sacrificing performance. These results demonstrate LoRA-XS's flexibility and scalability, making it well-suited for memory-constrained scenarios.}
    \centering
    \resizebox{0.5\textwidth}{!}{
\begin{tabular}{c|cccc}
\toprule
Model                       & Method                    & \makecell{\# Trainable \\ Parameters} & GSM8K & MATH \\
\hline
\multirow{5}{*}{\makecell{Mistral\\7B}} & Full FT                      & 7242M                   &  67.02     &   18.60   \\
                            & LoRA                        & 168M                    &  67.70     &   19.68  \\ \cline{2-5}
                            & \multirow{4}{*}{LoRA-XS}   & \textbf{3.67M}                        &  \textbf{70.35}     &   \textbf{20.96}   \\ 
                            &                             & 0.92M                        &  68.01     &   17.86   \\
                            &                             & 0.23M                        &  63.23     &   15.88   \\
                            &                             & 0.057M                       &  57.92     &   14.44   \\
\hline
\multirow{5}{*}{\makecell{Gemma\\7B}}   & Full FT                     & 8538M                   &  71.34     &   22.74   \\
                            & LoRA                        & 200M                    &  74.90     &   \textbf{31.28}  \\ \cline{2-5}
                            & \multirow{4}{*}{LoRA-XS}   & \textbf{3.21M}                  &  \textbf{76.42}     &   30.82   \\
                            &                             & 0.80M                   &  74.22     &   27.62   \\
                            &                             & 0.20M                   &  71.72     &   27.32   \\
                            &                            & 0.050M                  &  68.46     &   26.38  \\
\bottomrule 
\end{tabular}
}

    \label{tab:instruction_tuning}
\end{table}

The results of our instruction tuning experiments are summarized in~\Cref{tab:commonsense} and~\Cref{tab:instruction_tuning}.

In~\Cref{tab:commonsense}, we compare LoRA-XS with LoRA across eight commonsense reasoning datasets using LLaMA2-7B and LLaMA3-8B models. LoRA-XS consistently outperforms LoRA while using only a small fraction of its trainable parameters. For example, on the LLaMA2-7B model, LoRA-XS with just 3.67M trainable parameters achieves a significantly higher average accuracy than LoRA with 56M parameters, demonstrating exceptional parameter efficiency and scalability to billion-scale models.

In~\Cref{tab:instruction_tuning}, we evaluate LoRA-XS on mathematical reasoning tasks (GSM8K and MATH datasets) using Mistral-7B and Gemma-7B models. LoRA-XS not only performs competitively but also surpasses both LoRA and full fine-tuning in most cases.  Notably, on GSM8K, LoRA-XS outperforms LoRA by over 2 percentage points while using only 3.21M trainable parameters, compared to LoRA’s >150M parameters models.

\section{Ablation}
\label{sec:mainablation}
In this section, we present ablation experiments to better understand the efficiency of LoRA-XS, demonstrate our theoretical derivations in practice, and examine the role of singular vectors in transformer weights and adaptation layers.

\paragraph{Importance of Singular Vectors in Transformer Weights} We begin by analyzing the significance of singular vectors in the weight matrices of transformer models. Detailed results can be found in~\Cref{appendix:ablation_svd_1}. By examining different subsets of singular vectors (top, middle, and bottom) for various transformer weights (e.g., attention and feedforward modules), we find that the top singular vectors retain the most task-relevant knowledge. In contrast, the middle and bottom singular vectors contribute less to task performance, suggesting that they encode more subtle or less critical information. This observation aligns with our proposal to initialize LoRA-XS using top singular vectors.

\paragraph{Delta Weight Approximation and Singular Subspace Retention} We evaluate how accurately the full weight update $\Delta W$, obtained during fine-tuning, can be approximated by projecting it onto different subspaces of singular vectors from the SVD of the original pre-trained weight matrix $W$. Specifically, we experiment with retaining varying fractions of top, middle, and bottom singular vectors and assess their impact on downstream task performance. As summarized in~\Cref{appendix:ablation_svdsubspace}, the self-attention modules (query, key, value, attention output) exhibit minimal performance degradation when only 1\% or 10\% of singular vectors are retained, regardless of whether they come from the top or bottom subspaces. In contrast, output dense layers are more sensitive to these approximations and require a higher fraction of singular vectors to preserve accuracy. These results suggest that while self-attention layers can tolerate significant dimensionality reduction, output dense layers benefit from retaining a larger portion of the singular spectrum.

\paragraph{LoRA-XS Initialization} The initialization of matrices $A$ and $B$ is a key factor in the performance of LoRA-XS. We investigate three initialization strategies: random initialization, SVD of random matrices (SVD of random), and SVD of pre-trained weights (SVD of $W$). Please refer to~\Cref{appendix:ablation_loraxs_init} for the details. 

\begin{table}[ht!]
\caption{Performance of LoRA-XS under various initialization schemes. We present the best median scores across different learning rates for 5 seeds at rank $4$. We report Matthew’s correlation for CoLA and accuracy for the other tasks. Initializing LoRA-XS using the SVD of pre-trained weights (SVD of $W$) outperforms other methods on most tasks. For SST-2, the SVD of random matrices achieves a slightly higher score, likely due to the task's unique characteristics. Please refer to \Cref{appendix:ablation_loraxs_init} for further details.}
\centering
\resizebox{0.5\textwidth}{!}{
\begin{tabular}{c|c|c|c|c}
\hline
Init. Type & SST-2 & COLA & MRPC & QNLI \\
\hline
random & 94.72 & 58.53 & 85.78 & 88.80 \\
SVD of random & \textbf{94.84} & 55.27 & 84.31 & 88.34 \\
SVD of W & 94.72 & \textbf{60.11} & \textbf{87.50} & \textbf{90.94} \\

\hline
\end{tabular}
}
\label{main:ablation:svdinit}
\end{table}

As summarized in \Cref{main:ablation:svdinit}, our experiments show that using SVD on the pre-trained weight matrices generally leads to superior performance. This observation aligns with our theoretical framework, which claims that, assuming the considered task is similar to the task used for pre-training, SVD of the original weight matrix is the most effective initialization choice (see \Cref{theory}).

An exception to this trend is observed in the SST-2 task, where SVD of random matrices slightly outperforms SVD of $W$. We hypothesize that this is due to SST-2 being a sentiment classification task, which may not align as closely with the pre-training objective of language modeling as other tasks such as MRPC, CoLA, and QNLI. This insight reinforces our theoretical analysis, which suggests that SVD of pre-trained weights is most advantageous when the fine-tuning task shares similarities with the pre-training objective. Importantly, our empirical results show that LoRA-XS remains competitive across a wide range of tasks, including those less aligned with the pre-training objective (see~\Cref{tab:gluebenchmark}).

Additionally, we show that initializing \our{} with SVD of the pretrained weights accelerates convergence in the early stages of \our{} training (see \Cref{tab:ablationsvdvsnosvd}). This early advantage sets LoRA-XS apart from other ultra-efficient adaptation techniques, such as soft prompt tuning~\citep{lester2021power, li2021prefix}, which often exhibit slower convergence. By initializing $A$ and $B$ with information derived from the pre-trained model, LoRA-XS benefits from a more informed starting point, leading to more efficient and effective training.

\paragraph{Top vs. Bottom Singular Vector Initialization} In \Cref{ablation:topvsbottomsingularvectorsinit}, we further analyze whether it is more effective to initialize LoRA-XS with top or bottom singular vectors. Our analysis indicates that retaining the top singular vectors consistently yields better performance across various tasks.

\paragraph{Including Singular Values in Initialization} Lastly, in \Cref{ablation:sigmaornotinit}, we evaluate whether including singular values $\Sigma$ in the initialization of matrix $A$ enhances the performance of LoRA-XS. The results indicate improved performance when $\Sigma$ is included in most cases, suggesting that while singular values do not alter the direction of the corresponding singular vectors, they may play a crucial role in scaling and emphasizing their significance. However, one task performed better without $\Sigma$, which may suggest that certain types of tasks could benefit from a different approach to initialization.

\section{Conclusion}
\label{conclusion}

We introduce LoRA-XS, a parameter-efficient fine-tuning method that significantly reduces trainable parameters compared to existing approaches. LoRA-XS leverages low-rank adaptation with SVD to align adaptation matrices with the principal components of pre-trained weights. Our extensive experiments demonstrate its superior parameter efficiency while maintaining or surpassing accuracy. With strong theoretical foundations and proven empirical effectiveness, LoRA-XS stands as both a powerful fine-tuning method and a promising foundation for future advancements in efficient adaptation for large-scale models.

\subsubsection*{Acknowledgments}
The work of Klaudia Bałazy was supported by the National Centre of Science (Poland) Grant No. 2020/39/D/ST6/01332. Klaudia Bałazy is affiliated with Doctoral School of Exact and Natural Sciences at the Jagiellonian University. This research was partially funded by the National Science Centre, Poland, Grant No. 2023/49/B/ST6/01137 (work by Jacek Tabor). Some experiments were performed on servers purchased with funds from the flagship project entitled “Artificial Intelligence Computing Center Core Facility” from the DigiWorld Priority Research Area within the Excellence Initiative – Research University program at Jagiellonian University in Krakow.

\bibliography{ourbib}

\appendix

\clearpage

\section{Appendix Overview}
\label{appendix:intro}

This appendix provides additional details and supplementary materials to complement the main text of our paper on LoRA-XS. For the reader’s convenience, we outline the contents of each appendix section below.

\paragraph{Orthogonal Projection onto the LoRA-XS Subspace}

In~\Cref{appendix:theory_proof}, we provide a proof complementing our theory decribed in~\Cref{theory}.

\paragraph{Efficiency of LoRA-XS}
In~\Cref{appendix:param_eff_loraxs}, we analyze and compare the parameter efficiency of LoRA-XS against both LoRA and VeRA, demonstrating its superior performance with fewer trainable parameters. Additionally, we examine the computational overhead of the SVD step as well as the runtime performance of LoRA-XS.

\paragraph{Experimental Setup Details}
In~\Cref{appendix:setup}, we provide details on our experimental setup, including hyperparameters, datasets, and implementation specifics.

\paragraph{Ablation Study on Singular Value Retention}
In~\Cref{appendix:ablation_svd_1}, we investigate the impact of retaining different subsets of singular vectors (top, middle, and bottom) in the transformer model on task performance, highlighting the importance of the top singular vectors.

\paragraph{Ablation Study on the Effect of Retaining Singular Vector Subspaces on Delta Weight Approximation}
In~\Cref{appendix:ablation_svdsubspace}, we evaluate how accurately the full weight update $\Delta W$ can be approximated by projecting it onto various singular vector subspaces, and we assess the sensitivity of different transformer modules to these approximations.

\paragraph{Ablation Study on LoRA-XS Initialization}
In~\Cref{appendix:ablation_loraxs_init}, we compare different initialization strategies for LoRA-XS (including random initialization, SVD of random matrices, and SVD of pre-trained weights) to determine their effects on performance.

\paragraph{Ablation Study on Top vs. Bottom Singular Vector Initialization}
In~\Cref{ablation:topvsbottomsingularvectorsinit}, we examine whether initializing LoRA-XS with top or bottom singular vectors is more effective, and our results indicate that the top singular vectors consistently yield better performance.

\paragraph{Ablation Study on the Importance of Singular Values for LoRA-XS}
In~\Cref{ablation:sigmaornotinit}, we explore whether including singular values in the initialization of LoRA-XS improves performance, and we discuss the conditions under which this approach is most beneficial.

\paragraph{Acknowledgments} We acknowledge the use of ChatGPT for grammar and spelling checking.

\clearpage

\section{Orthogonal Projection onto the LoRA-XS Subspace}

\label{appendix:theory_proof}

In this subsection, we derive the orthogonal projection onto the subspace $S^r_{A,B}$, which constrains fine-tuning updates in LoRA-XS. Given a full gradient update, our method efficiently projects it onto this subspace using a closed-form expression. This ensures that updates remain within the constrained manifold while preserving essential information. As a result, the number of trainable parameters can be flexibly adjusted, ranging from a single parameter to an $r \times r$ matrix, depending on the chosen rank.

Given fixed orthogonal matrices $A \in \R^{r \times n} ,B \in \R^{n \times r}$, recall that 
$$
S^r_{A,B}=\{AXB^T:X \in \R^{r \times r}\}.
$$

We show that one can easily compute the orthogonal projection on $S^r_{A,B}$. Namely,
$$
p_{A,B}(X)=A[A^TXB]B^T \text{ for }X \in \R^{n \times n},
$$
is the orthogonal projection with respect to Frobenius scalar product in the space of matrices on $S^r_{A,B}$.

\begin{proof}
To prove the above, let us recall that $p$ is an orthogonal projection iff $p^2=p$ and $p=p^T$. To check if it is projection onto space $S$ we have to additionally verify if $p(x) \in S$ for arbitrary $x$ and $p(x)=x$ for $x \in S$.

Let us first check that $p^2_{A,B}=p_{A,B}$:
\begin{multline}
p^2_{A,B}(X)=A(A^TA)A^TXB(B^TB)B^T=\\
AA^TXBB^T=p_{A,B}(X).
\end{multline}

Now we check if $p_{A,B}=p_{A,B}^T$. Since for a self adjoint map $C$ we have $\il{Cx,y}=\il{x,CY}$, and $AA^T,BB^T$ are self-adjoint, we get:
\begin{multline}
\il{p_{A,B}X,Y}=\il{AA^TXBB^T,Y}=\\
\tr{(AA^TXBB^T)^TY}=\tr{BB^TX^TAA^TY}=\\
\tr{AA^TYBB^TX^T}=\il{p_{A,B}Y,X}.
\end{multline}

Clearly, directly from definition $p_{A,B}(X)=A[A^TXB]B^T \in S^r_{A,B}$ for an arbitrary $X \in \R^{n \times n}$.
Finally, we check that $p_{A,B}(W)=W$ for $W \in S^r_{A,B}$. Since $W \in S^r_{A,B}$, $W=AXB^T$ for some $X$. Consequently,
$$
p_{A,B}(W)=AA^T(AXB^T)BB^T=AXB^T=W.
$$

Thus we have show that the family $S^r_{A,B}$ of $r^2$-dimensional subspaces of $\R^{n \times n}$ allows an easy formula for orthogonal projection.

\end{proof}

\clearpage

\section{Efficiency of LoRA-XS}
\label{appendix:param_eff_loraxs}

\subsection{Parameter Efficiency}

We make the following observation on the parameter efficiency of LoRA-XS compared to LoRA and VeRA methods.\footnote{It is worth noting that in these calculations we only consider the additional LoRA/VeRA/LoRA-XS parameters, and for encoder-only models, additional trainable parameters such as classifiers parameters may be added. However, these are common to all approaches and thus do not influence our comparative calculations.}

\textbf{Observation:} LoRA-XS demonstrates superior parameter efficiency compared to both LoRA and VeRA.

For simplicity, let's consider a transformer model with $L$ fine-tuned layers, each consisting of $q$ number of $W \in \mathbb{R}^{n \times n}$ matrices.

For LoRA, the number of trainable parameters is given by:
\begin{equation}
    P_{\text{LoRA}} = L \times q \times r \times 2n,
\end{equation}

For VeRA, the number of trainable parameters is given by:
\begin{equation}
    P_{\text{VeRA}} = L \times q \times (n + r),
\end{equation}

For LoRA-XS, the number of trainable parameters is given by:
\begin{equation}
    P_{\text{LoRA-XS}} = L \times q \times r^2.
\end{equation}

To compare the parameter efficiency, we compute the ratios of the number of trainable parameters between the methods. The ratio of trainable parameters for LoRA to LoRA-XS is:
\begin{equation}
    \frac{P_{\text{LoRA}}}{P_{\text{LoRA-XS}}} = \frac{L \times q \times r \times 2n}{L \times q \times r^2} = \frac{2n}{r},
\end{equation}

Similarly, the ratio of trainable parameters for VeRA to LoRA-XS is:
\begin{equation}
    \frac{P_{\text{VeRA}}}{P_{\text{LoRA-XS}}} = \frac{L \times q \times (n + r)}{L \times q \times r^2} = \frac{n + r}{r^2}.
\end{equation}

As the model dimension $n$ becomes very large compared to the rank $r$, the benefit of LoRA-XS over LoRA and VeRA becomes more pronounced. Specifically for large $n$:
\begin{equation}
    \frac{P_{\text{LoRA}}}{P_{\text{LoRA-XS}}} \approx \frac{2n}{r}
    \quad \text{and} \quad
    \frac{P_{\text{VeRA}}}{P_{\text{LoRA-XS}}} \approx \frac{n}{r^2}.
\end{equation}

This indicates that for large models, LoRA and VeRA require significantly more parameters than LoRA-XS, with the difference growing linearly with $n$ (\ie, model's hidden dimension). This makes LoRA-XS especially suitable for fine-tuning large language models where parameter efficiency is crucial.

To provide an example, for RoBERTa-large~\citep{liu2019roberta}, which consists of 24 layers, assuming $q = 2$ (two additional trainable modules per layer), with each $W$ matrix of size $1024 \times 1024$ and $r = 16$, the number of trainable parameters for each method is as follows: For LoRA, $P_{\text{LoRA}} = 1,572,864$; for VeRA, $P_{\text{VeRA}} = 50,400$; for LoRA-XS, $P_{\text{LoRA-XS}} = 12,288$. Thus, LoRA requires about 31.2 times more parameters than VeRA and 128 times more than LoRA-XS, while VeRA requires about 4 times more parameters than LoRA-XS, highlighting the substantial parameter efficiency of LoRA-XS.

\subsection{Computational Cost of SVD Initialization}

It is noteworthy that the SVD computation used in LoRA-XS is a one-time operation for each weight matrix, which is considerably faster than full fine-tuning and is offset by the efficiency gains achieved during adaptation.

To quantify the computational cost of SVD, we performed timing experiments using RoBERTa-large on H100 GPU for the SST-2 and MRPC tasks. We compared the total fine-tuning time with the SVD initialization time for all matrices on SST-2 (20 epochs) and MRPC (50 epochs) with ranks of 4 and 25. These results demonstrate that the SVD cost is minimal relative to the overall fine-tuning time, confirming that the initialization step does not significantly affect computational efficiency. Table~\ref{tab:svd_timing} shows that SVD initialization constitutes less than 1\% of the total fine-tuning time.

\begin{table}[h]
\centering
\caption{SVD initialization cost relative to total fine-tuning time for RoBERTa-large on SST-2 and MRPC tasks.}
\label{tab:svd_timing}
\begin{tabular}{l|c|c|c|c}
\hline
\textbf{Task} & \textbf{Rank} & \textbf{\makecell{Total Fine-Tuning\\ Time (s)}} & \textbf{SVD Init (s)} & \textbf{\% Overhead} \\
\hline
SST-2 & 4  & 7310 & 10.6 & 0.14\% \\
SST-2 & 25 & 6980 & 19.1 & 0.27\% \\
MRPC  & 4  & 1215 & 10.8 & 0.89\% \\
MRPC  & 25 & 3104 & 18.9 & 0.61\% \\
\hline
\end{tabular}
\end{table}

\subsection{Runtime}

To ensure LoRA-XS’s efficiency in practice, we conducted runtime measurements on RoBERTa-large (MRPC, 10 epochs) using an H100 GPU. As shown in Table~\ref{tab:runtime_comparison}, the runtime difference between LoRA-XS and LoRA is minimal across different ranks, demonstrating that LoRA-XS introduces negligible overhead.

\begin{table}[h]
\centering
\caption{Runtime comparison (in seconds) between LoRA and LoRA-XS on RoBERTa-large (MRPC task, 10 epochs).}
\label{tab:runtime_comparison}
\begin{tabular}{c|c|c}
\hline
\textbf{Model} & \textbf{Rank} & \textbf{Runtime (s)} \\
\hline
LoRA      & 16 & 198.8 \\
LoRA-XS   & 16 & 200.7 \\
LoRA      & 32 & 200.6 \\
LoRA-XS   & 32 & 201.7 \\
LoRA      & 64 & 204.5 \\
LoRA-XS   & 64 & 204.8 \\
\hline
\end{tabular}
\end{table}

\clearpage

\section{Experimental Setup Details}
\label{appendix:setup}

In this section, we provide detailed information on the experimental setup and hyperparameters used in our experiments.

As mentioned in~\Cref{experiments}, for our main experiments, each LoRA-XS module is initialized using Singular Value Decomposition (SVD) of the corresponding pre-trained weight matrix $W$. The initialization process involves using truncated SVD~\citep{halko2011finding}. We take LoRA and VeRA scores from their respective papers~\citep{lora,kopiczko2023vera}.

For all our experiments and the ablation study, models are trained with the AdamW optimizer~\citep{loshchilov2017decoupled}, following methodologies from LoRA and VERA~\citep{lora,kopiczko2023vera}. We utilize the HuggingFace Transformers library~\citep{wolf2019huggingface} for Transformer-based models~\citep{vaswani2017attention} and implement our LoRA-XS method on top of the Huggingface PEFT repository~\citep{hfpeft}. Detailed setups and hyperparameters for each experiment are outlined in the following subsections.

Our preliminary findings showed that when the number of trainable parameters is limited, using a lower rank $r$ with more LoRA-XS modules yields better results. This guided our strategy of using smaller ranks while distributing more LoRA-XS modules than LoRA and VeRA, which mainly added adaptation modules to the Query and Value matrices in the main experiments. By spreading LoRA-XS across additional components and keeping the rank low, we achieve a balanced parameter allocation without significantly increasing the number of trainable parameters.

\subsection{GLUE Benchmark}
\label{appendix:setup_glue}

For the GLUE Benchmark experiments, we integrate LoRA-XS modules into the Query ($W_q$), Value ($W_v$), Attention Output ($W_o$), and first Fully Connected ($FC_1$) weight matrices of the transformer model~\citep{vaswani2017attention}. RoBERTa-large~\citep{liu2019roberta} serves as the base model. LoRA-XS's rank, $r$, is varied between 4 and 25, corresponding to 16 to 625 trainable parameters per module. Following previous work~\citep{lora}, LoRA modules for MRPC, RTE, and STS-B are initialized with weights fine-tuned on the MNLI task. As in VERA~\cite{kopiczko2023vera}, larger datasets like MNLI and QQP are excluded from our experiments due to computational constraints.

Hyperparameters were selected via grid search, and the chosen values are detailed in~\Cref{tab:gluehyperparams}. The sequence length is set to 128, with a warm-up ratio of 0.06. All tasks used a batch size of 32 and were trained on a single A100 40GB GPU, GeForce RTX 4090, or H100 80GB, depending on resource availability. We fix the LoRA-XS scaling factor $\alpha$ to 16.

\begin{table}[h!]
    \small
    \centering
    \vskip 0.1in
    \begin{tabular}{@{}c|c|c|c|c@{}}
         \midrule
         Task & Rank & LoRA-XS LR & Classifier LR & Epochs \\
         \midrule

        \multirow{6}{*}{SST-2}
        & 4 & 1E-3 & 1E-3 & 20 \\
        & 8 & 1E-3 & 1E-3 & 20 \\
        & 12 & 5E-3 & 1E-3 & 20 \\
        & 16 & 1E-3 & 5E-4 & 20 \\
        & 20 & 1E-3 & 5E-3 & 20 \\
        & 25 & 2E-3 & 1E-3 & 20 \\
        \midrule

        \multirow{6}{*}{MRPC}
        & 4 & 1E-3 & 1E-3 & 50 \\
        & 8 & 1E-3 & 6E-4 & 50 \\
        & 12 & 1E-3 & 1E-3 & 50 \\
        & 16 & 1E-3 & 6E-4 & 50 \\
        & 20 & 1E-3 & 6E-4 & 50 \\
        & 25 & 1E-3 & 6E-4 & 50 \\
        \midrule

        \multirow{6}{*}{CoLA}
        & 4 & 1E-3 & 5E-3 & 50 \\
        & 8 & 1E-3 & 5E-3 & 50 \\
        & 12 & 1E-3 & 5E-3 & 50 \\
        & 16 & 1E-3 & 1E-2 & 50 \\
        & 20 & 1E-3 & 5E-3 & 50 \\
        & 25 & 1E-3 & 5E-3 & 50 \\
        \midrule

        \multirow{6}{*}{QNLI}
        & 4 & 1E-3 & 5E-4 & 10 \\
        & 8 & 1E-3 & 1E-3 & 10 \\
        & 12 & 1E-3 & 5E-4 & 10 \\
        & 16 & 1E-3 & 1E-3 & 15 \\
        & 20 & 1E-3 & 5E-4 & 10 \\
        & 25 & 2E-3 & 6E-4 & 15 \\
        \midrule

        \multirow{6}{*}{RTE}
        & 4 & 1E-3 & 6E-4 & 50 \\
        & 8 & 6E-4 & 6E-4 & 50 \\
        & 12 & 1E-3 & 6E-4 & 50 \\
        & 16 & 1E-3 & 1E-3 & 50 \\
        & 20 & 1E-3 & 1E-3 & 50 \\
        & 25 & 1E-3 & 6E-4 & 50 \\
        \midrule

        \multirow{6}{*}{STS-B}
        & 4 & 6E-4 & 6E-4 & 50 \\
        & 8 & 1E-3 & 1E-3 & 50 \\
        & 12 & 1E-3 & 1E-3 & 50 \\
        & 16 & 1E-3 & 1E-3 & 50 \\
        & 20 & 1E-3 & 6E-4 & 50 \\
        & 25 & 1E-3 & 6E-4 & 50 \\
         \bottomrule
    \end{tabular}
    \vspace{10pt}
    \caption{Hyperparameters selected via grid search for RoBERTa-large fine-tuned with LoRA-XS across various GLUE tasks. The reported hyperparameters reflect the best-performing settings found during the search.}
    \label{tab:gluehyperparams}
\end{table}

\subsection{Instruction Tuning Experiments}
\label{appendix:setup_instruction}

As mentioned earlier, for all instruction tuning experiments, LoRA-XS modules are added to the Query, Key, Value, Attention Output, and all three fully connected weight matrices. Moreover, during training, the cross-entropy loss is calculated only on the generated tokens (\ie~ not considering instruction tokens).

For mathematical reasoning tasks, we perform the instruction tuning experiments on Mistral-7B-v0.1~\citep{jiang2023mistral} and Gemma-7B~\citep{team2024gemma} decoder-only models. We use a batch size of 128, the AdamW optimizer~\citep{loshchilov2017decoupled}, and train on 100k samples of the MetaMathQA dataset. For ranks 16, 32, and 64, we train for 2 epochs with a learning rate of $4e-3$. For a larger rank of 128, we train for 3 epochs with a reduced learning rate of $7e-4$ to ensure adequate convergence for the increased number of trainable parameters while minimizing overfitting. Interestingly, when evaluating some earlier checkpoints (prior to the end of training), we observed validation results that exceeded those that we reported, suggesting that early stopping could further enhance LoRA-XS's performance. Models are evaluated on the GSM8K and MATH datasets. A warm-up ratio of 0.02 and a cosine learning rate scheduler are used. The LoRA-XS parameter $\alpha$ equals the rank.

For commonsense reasoning, the experiments are conducted on LLaMA2 7B and LLaMA3 8B decoder-only models. We train the models for 3 epochs on a collection of 170K commonsense samples~\citep{hu2023llm}. Models are then evaluated on BoolQ~\citep{clark2019boolq}, PIQA~\citep{bisk2020piqa}, SIQA~\citep{sap2019socialiqa}, HellaSwag~\citep{zellers2019hellaswag}, WinoGrande~\citep{sakaguchi2021winogrande}, OBQA~\citep{mihaylov2018can}, ARC-c (challenge) and ARC-e (easy) datasets~\citep{clark2018think}.
We use a batch size of 64 and the learning rate is set to $10^{-3}$ with the AdamW optimizer. The warmup is for 100 steps, and a linear decay is used for the learning rate scheduler. The LoRA-XS parameter $\alpha$ is set to 64.

In both mathematical and commonsense reasoning experiments, LoRA-XS modules are added to the key, query, value, attention output, up projection, down projection, and gate projection layers. Two A100 80GB GPUs or H100 80GB GPUs were used for fine-tuning.

\subsection{Ablation experimental setup}

All ablation experiments were conducted on the GLUE benchmark using RoBERTa-large as the base model. Due to computational limits, ablations were performed on a subset of tasks. Results are reported as median values across 5 seeds, with the best model selected based on validation performance. Different learning rates were applied as detailed in the respective sections.

Training was performed on a single A100 40GB GPU or H100 80GB GPU with a batch size of 32. We used a fixed $\alpha$ of 16, a sequence length of 128, and a warm-up ratio of 0.06. Note that MNLI initialization was not used for ablation studies.

\clearpage

\section{Ablation Study: Singular Value Retention}
\label{appendix:ablation_svd_1}
To gain insights into the importance of singular vectors in the pre-trained model's performance, we conduct an ablation study on fine-tuned RoBERTa-large model by performing Singular Value Decomposition (SVD) on selected weight matrices. Specifically, we decompose the matrices into their singular values and vectors, retaining only a fraction \( r_{\text{frac}} \) of the singular values: either from the \textbf{top}, \textbf{middle}, or \textbf{bottom} of the spectrum, while zeroing out the remaining ones. This experiment aims to explore which portions of the singular value spectrum contribute most to the model's performance.

The experiment is carried out on RoBERTa-large fine-tuned on MRPC, SST-2, and MNLI tasks using default hyperparameters, including a learning rate of 2e-5 and 5 epochs for MRPC, 3 epochs for SST-2, and MNLI. We evaluate the performance of various weight matrices within the transformer layers, such as `query`, `value`, `key`, `attention.output.dense`, `intermediate.dense`, and `output.dense`. Each matrix is tested across the following configurations:
\begin{itemize}
    \item \textbf{Top-r}: Retaining the top \( r_{\text{frac}} \) singular values.
    \item \textbf{Middle-r}: Retaining the middle \( r_{\text{frac}} \) singular values.
    \item \textbf{Bottom-r}: Retaining the bottom \( r_{\text{frac}} \) singular values.
\end{itemize}

The results, shown in Tables \ref{tab:mrpc_results}, \ref{tab:sst2_results}, and \ref{tab:mnli_results}, reveal several key insights. First, retaining only the top singular values consistently yields strong performance, particularly in the `query`, `value`, and `key` matrices. Interestingly, retaining a small fraction (\( r_{\text{frac}} = 0.1 \)) of the top singular values still achieves reasonable accuracy in tasks like SST-2 and MNLI. In contrast, retaining the middle or bottom singular values generally leads to a sharp performance degradation, suggesting their limited role in maintaining task-specific knowledge.

An exception to this pattern is observed in the `intermediate.dense` matrix, where preserving the bottom singular values for MRPC and SST-2 yields better performance than retaining the top or middle singular values. This suggests that the `intermediate.dense` matrix may store more task-specific information in the lower-ranked singular vectors. One possible explanation is that intermediate layers could be more sensitive to the distribution of information, requiring a broader spread across singular values, or that they store certain nuanced representations directly in the lower spectrum.

These findings provide initial evidence that the top singular vectors capture the most essential information in transformer weights. This reinforces the intuition that low-rank adaptations, such as those employed in LoRA-XS, can be highly effective in parameter-efficient fine-tuning scenarios.

\begin{table*}[ht]
\centering
\small
\begin{tabular}{|c|c|c|c|c|}
\hline
$r_{\text{frac}}$  & Module & Top $r_{\text{frac}}$ Acc & Middle $r_{\text{frac}}$  Acc & Bottom $r_{\text{frac}}$  Acc \\
\hline
0.0 & query & 31.62 & 31.62 & 31.62 \\
0.0 & value & 31.62 & 31.62 & 31.62 \\
0.0 & key & 31.62 & 31.62 & 31.62 \\
0.0 & attention.output.dense & 31.62 & 31.62 & 31.62 \\
0.0 & intermediate.dense & 31.62 & 31.62 & 31.62 \\
0.0 & output.dense & 31.62 & 31.62 & 31.62 \\
\midrule
0.1 & query & 70.83 & 31.62 & 31.62 \\
0.1 & value & 69.61 & 31.62 & 31.62 \\
0.1 & key & 55.39 & 31.62 & 31.62 \\
0.1 & attention.output.dense & 31.86 & 31.62 & 31.62 \\
0.1 & intermediate.dense & 31.86 & 31.62 & 68.38 \\
0.1 & output.dense & 31.86 & 31.62 & 31.62 \\
\midrule
0.25 & query & 88.24 & 31.62 & 31.62 \\
0.25 & value & 77.21 & 33.58 & 31.62 \\
0.25 & key & 87.99 & 31.62 & 31.62 \\
0.25 & attention.output.dense & 86.76 & 31.86 & 31.62 \\
0.25 & intermediate.dense & 33.58 & 31.62 & 68.38 \\
0.25 & output.dense & 32.11 & 31.62 & 31.62 \\
\midrule
0.5 & query & 88.24 & 31.62 & 31.62 \\
0.5 & value & 85.05 & 31.37 & 31.62 \\
0.5 & key & 88.97 & 31.62 & 31.62 \\
0.5 & attention.output.dense & 88.97 & 32.60 & 31.62 \\
0.5 & intermediate.dense & 31.62 & 31.62 & 68.38 \\
0.5 & output.dense & 83.82 & 31.62 & 31.62 \\
\midrule
0.75 & query & 88.97 & 31.86 & 31.62 \\
0.75 & value & 88.48 & 31.62 & 31.62 \\
0.75 & key & 88.97 & 31.62 & 31.62 \\
0.75 & attention.output.dense & 88.73 & 65.20 & 32.11 \\
0.75 & intermediate.dense & 35.29 & 31.62 & 31.62 \\
0.75 & output.dense & 86.52 & 68.38 & 31.62 \\
\midrule
0.9 & query & 88.48 & 35.78 & 32.11 \\
0.9 & value & 88.24 & 42.89 & 31.62 \\
0.9 & key & 88.73 & 31.86 & 31.62 \\
0.9 & attention.output.dense & 88.73 & 84.31 & 67.65 \\
0.9 & intermediate.dense & 37.25 & 31.62 & 68.38 \\
0.9 & output.dense & 88.24 & 68.63 & 68.38 \\
\midrule
1.0 & query & 88.73 & 88.73 & 88.73 \\
1.0 & value & 88.73 & 88.73 & 88.73 \\
1.0 & key & 88.73 & 88.73 & 88.73 \\
1.0 & attention.output.dense & 88.73 & 88.73 & 88.73 \\
1.0 & intermediate.dense & 88.73 & 88.73 & 88.73 \\
1.0 & output.dense & 88.73 & 88.73 & 88.73 \\
\midrule
\end{tabular}
\caption{Performance on the MRPC task using varying fractions of singular values (Top, Middle, and Bottom) retained across different fine-tuned RoBERTa-large weight matrices.}
\label{tab:mrpc_results}
\end{table*}

\begin{table*}[ht]
\centering
\small
\begin{tabular}{|c|c|c|c|c|}
\hline
$r_{\text{frac}}$  & Module & Top $r_{\text{frac}}$ Acc & Middle $r_{\text{frac}}$  Acc & Bottom $r_{\text{frac}}$  Acc \\
\hline
0.0 & query & 50.92 & 50.92 & 50.92 \\
0.0 & value & 50.92 & 50.92 & 50.92 \\
0.0 & key & 50.92 & 50.92 & 50.92 \\
0.0 & attention.output.dense & 50.92 & 50.92 & 50.92 \\
0.0 & intermediate.dense & 50.92 & 50.92 & 50.92 \\
0.0 & output.dense & 49.08 & 49.08 & 49.08 \\
\midrule
0.1 & query & 93.81 & 50.92 & 50.92 \\
0.1 & value & 91.74 & 51.26 & 49.89 \\
0.1 & key & 93.92 & 50.92 & 50.92 \\
0.1 & attention.output.dense & 90.60 & 55.50 & 50.92 \\
0.1 & intermediate.dense & 50.92 & 52.18 & 50.92 \\
0.1 & output.dense & 51.03 & 49.08 & 49.08 \\
\midrule
0.25 & query & 95.30 & 50.92 & 50.92 \\
0.25 & value & 94.84 & 58.72 & 50.92 \\
0.25 & key & 95.64 & 51.03 & 50.92 \\
0.25 & attention.output.dense & 95.53 & 61.12 & 50.57 \\
0.25 & intermediate.dense & 50.92 & 49.89 & 50.92 \\
0.25 & output.dense & 63.19 & 53.90 & 49.08 \\
\midrule
0.5 & query & 95.64 & 51.38 & 50.92 \\
0.5 & value & 95.76 & 53.10 & 54.70 \\
0.5 & key & 95.64 & 51.61 & 50.92 \\
0.5 & attention.output.dense & 95.64 & 78.78 & 63.65 \\
0.5 & intermediate.dense & 60.55 & 49.08 & 50.92 \\
0.5 & output.dense & 93.00 & 49.31 & 49.08 \\
\midrule
0.75 & query & 95.76 & 53.10 & 51.49 \\
0.75 & value & 95.64 & 67.55 & 52.41 \\
0.75 & key & 95.76 & 54.36 & 51.49 \\
0.75 & attention.output.dense & 95.87 & 91.28 & 80.16 \\
0.75 & intermediate.dense & 86.81 & 49.08 & 50.92 \\
0.75 & output.dense & 95.76 & 50.92 & 49.08 \\
\midrule
0.9 & query & 95.64 & 62.04 & 55.05 \\
0.9 & value & 95.87 & 86.93 & 73.05 \\
0.9 & key & 95.87 & 61.58 & 55.50 \\
0.9 & attention.output.dense & 95.87 & 94.38 & 92.89 \\
0.9 & intermediate.dense & 89.91 & 49.08 & 49.20 \\
0.9 & output.dense & 95.76 & 50.92 & 50.92 \\
\midrule
1.0 & query & 95.87 & 95.87 & 95.87 \\
1.0 & value & 95.87 & 95.87 & 95.87 \\
1.0 & key & 95.87 & 95.87 & 95.87 \\
1.0 & attention.output.dense & 95.87 & 95.87 & 95.87 \\
1.0 & intermediate.dense & 95.87 & 95.87 & 95.87 \\
1.0 & output.dense & 95.87 & 95.87 & 95.87 \\
\midrule
\end{tabular}
\caption{Performance on the SST2 task using varying fractions of singular values (Top, Middle, and Bottom) retained across different fine-tuned RoBERTa-large weight matrices.}
\label{tab:sst2_results}
\end{table*}

\begin{table*}[ht]
\centering
\small
\begin{tabular}{|c|c|c|c|c|}
\hline
$r_{\text{frac}}$  & Module & Top $r_{\text{frac}}$ Acc & Middle $r_{\text{frac}}$  Acc & Bottom $r_{\text{frac}}$  Acc \\
\hline
0.0 & query & 33.88 & 33.88 & 33.88 \\
0.0 & value & 31.82 & 31.82 & 31.82 \\
0.0 & key & 34.82 & 34.82 & 34.82 \\
0.0 & attention.output.dense & 31.82 & 31.82 & 31.82 \\
0.0 & intermediate.dense & 32.74 & 32.74 & 32.74 \\
0.0 & output.dense & 31.82 & 31.82 & 31.82 \\
\midrule
0.1 & query & 68.28 & 34.19 & 34.03 \\
0.1 & value & 81.65 & 31.82 & 31.82 \\
0.1 & key & 70.58 & 34.49 & 34.95 \\
0.1 & attention.output.dense & 70.19 & 31.82 & 31.82 \\
0.1 & intermediate.dense & 33.40 & 34.29 & 33.68 \\
0.1 & output.dense & 31.84 & 31.82 & 31.82 \\
\midrule
0.25 & query & 86.87 & 34.35 & 33.66 \\
0.25 & value & 89.00 & 31.86 & 31.82 \\
0.25 & key & 88.39 & 34.67 & 35.10 \\
0.25 & attention.output.dense & 88.82 & 31.82 & 31.82 \\
0.25 & intermediate.dense & 32.30 & 31.82 & 35.45 \\
0.25 & output.dense & 35.62 & 31.82 & 31.82 \\
\midrule
0.5 & query & 90.07 & 34.14 & 33.83 \\
0.5 & value & 89.79 & 31.82 & 31.80 \\
0.5 & key & 90.22 & 35.03 & 34.98 \\
0.5 & attention.output.dense & 90.02 & 31.78 & 31.82 \\
0.5 & intermediate.dense & 31.85 & 32.74 & 35.45 \\
0.5 & output.dense & 80.23 & 31.75 & 31.82 \\
\midrule
0.75 & query & 90.14 & 33.71 & 33.74 \\
0.75 & value & 90.21 & 38.03 & 31.82 \\
0.75 & key & 90.31 & 36.72 & 35.01 \\
0.75 & attention.output.dense & 90.29 & 51.61 & 31.79 \\
0.75 & intermediate.dense & 36.79 & 35.45 & 35.45 \\
0.75 & output.dense & 89.48 & 32.33 & 32.28 \\
\midrule
0.9 & query & 90.38 & 43.21 & 34.44 \\
0.9 & value & 90.45 & 71.56 & 41.43 \\
0.9 & key & 90.29 & 39.03 & 37.56 \\
0.9 & attention.output.dense & 90.30 & 83.49 & 65.65 \\
0.9 & intermediate.dense & 38.98 & 35.45 & 35.45 \\
0.9 & output.dense & 89.99 & 32.08 & 33.76 \\
\midrule
1.0 & query & 90.28 & 90.28 & 90.28 \\
1.0 & value & 90.28 & 90.28 & 90.28 \\
1.0 & key & 90.28 & 90.28 & 90.28 \\
1.0 & attention.output.dense & 90.28 & 90.28 & 90.28 \\
1.0 & intermediate.dense & 90.28 & 90.28 & 90.28 \\
1.0 & output.dense & 90.28 & 90.28 & 90.28 \\
\midrule
\end{tabular}
\caption{Performance on the MNLI task using varying fractions of singular values (Top, Middle, and Bottom) retained across different fine-tuned RoBERTa-large weight matrices.}
\label{tab:mnli_results}
\end{table*}

\clearpage

\section{Ablation Study: Effect of Retaining Singular Vector Subspaces on Delta Weight Approximation}
\label{appendix:ablation_svdsubspace}

In this ablation study, we assess how well $\Delta W$, representing the full weight updates, can be approximated by projecting it onto a subspace spanned by a subset of the singular vectors obtained from the SVD of the pre-trained model's weights $W$. Our goal is to evaluate how different subspaces (top, middle, or bottom) affect downstream performance across several GLUE tasks.

\subsection{Theoretical Framework}

The pre-trained model's weights $W$ are decomposed using singular value decomposition (SVD) as follows:

\[
W = U \Sigma V^T,
\]

where $U$ and $V$ are orthogonal matrices representing the left and right singular vectors, respectively, and $\Sigma$ is the diagonal matrix of singular values. During fine-tuning or LoRA adaptation, the model's weights are updated, resulting in a weight update $\Delta W$. This leads to the updated weight matrix being expressed as:

\[
W + \Delta W = U (\Sigma + C) V^T,
\]

where $\Delta W$ can be represented in the singular vector basis of $W$ as:

\[
\Delta W = U C V^T, \quad C = U^T \Delta W V.
\]

The matrix $C$ captures how the weight update $\Delta W$ projects onto the singular vector subspaces of $W$. We propose approximating $\Delta W$ by retaining only specific subspaces of $U$ and $V$, focusing on subspaces corresponding to the top, middle, or bottom singular vectors. This leads to the following approximation:

\[
\Delta W_{\text{approx}} = U C_r V^T,
\]

where $C_r$ is an $r \times r$ block matrix, constructed by selecting singular vector subspaces (top, middle, or bottom). By retaining different subspaces, we explore the sensitivity of the model’s performance to various parts of the singular spectrum.

\begin{table}[h]
\centering
\small
\begin{tabular}{|c|c|c|c|c|c|}
\hline
Module & $r_{\text{frac}}$ & retain & MRPC & SST-2 & MNLI \\
\hline
\multirow{12}{*}{query} 
 & 0.01 & top    & 89.71 & 95.64 & 90.08 \\ 
 & 0.01 & middle & 89.71 & 95.64 & 90.05 \\ 
 & 0.01 & bottom & 89.71 & 95.64 & 90.05 \\ 
 & 0.1  & top    & 89.71 & 95.64 & 90.07 \\ 
 & 0.1  & middle & 89.71 & 95.64 & 90.05 \\ 
 & 0.1  & bottom & 89.71 & 95.53 & 90.10 \\ 
 & 0.25 & top    & 89.71 & 95.64 & 90.14 \\ 
 & 0.25 & middle & 89.71 & 95.53 & 90.08 \\ 
 & 0.25 & bottom & 89.71 & 95.53 & 90.14 \\ 
 & 0.5  & top    & 89.46 & 95.64 & 90.20 \\ 
 & 0.5  & middle & 89.71 & 95.53 & 90.16 \\ 
 & 0.5  & bottom & 89.71 & 95.76 & 90.16 \\ 
\hline
\multirow{12}{*}{in.dense} 
 & 0.01 & top    & 88.97 & 95.53 & 88.16 \\ 
 & 0.01 & middle & 88.48 & 95.53 & 88.45 \\ 
 & 0.01 & bottom & 88.73 & 95.53 & 88.46 \\ 
 & 0.1  & top    & 88.73 & 95.41 & 88.16 \\ 
 & 0.1  & middle & 88.48 & 95.53 & 88.46 \\ 
 & 0.1  & bottom & 88.73 & 95.53 & 88.46 \\ 
 & 0.25 & top    & 89.22 & 95.41 & 88.18 \\ 
 & 0.25 & middle & 88.73 & 95.53 & 88.48 \\ 
 & 0.25 & bottom & 88.48 & 95.53 & 88.55 \\ 
 & 0.5  & top    & 89.22 & 95.41 & 88.40 \\ 
 & 0.5  & middle & 88.48 & 95.53 & 88.60 \\ 
 & 0.5  & bottom & 88.73 & 95.53 & 88.58 \\ 
\hline
\multirow{12}{*}{out.dense} 
 & 0.01 & top    & 79.41 & 94.04 & 86.82 \\ 
 & 0.01 & middle & 79.41 & 94.04 & 86.84 \\ 
 & 0.01 & bottom & 79.41 & 94.04 & 86.84 \\ 
 & 0.1  & top    & 80.15 & 94.61 & 87.05 \\ 
 & 0.1  & middle & 79.41 & 94.04 & 86.82 \\ 
 & 0.1  & bottom & 80.64 & 94.72 & 87.17 \\ 
 & 0.25 & top    & 81.37 & 95.18 & 88.76 \\ 
 & 0.25 & middle & 79.90 & 94.27 & 87.05 \\ 
 & 0.25 & bottom & 84.31 & 95.53 & 88.26 \\ 
 & 0.5  & top    & 86.27 & 95.53 & 89.34 \\ 
 & 0.5  & middle & 80.88 & 94.61 & 87.72 \\ 
 & 0.5  & bottom & 87.75 & 95.87 & 89.01 \\ 
\hline
\end{tabular}
\caption{Performance results of the SVD-based delta weight approximation experiment across different tasks (MRPC, SST-2, MNLI) and modules (query, intermediate dense (in.dense) and output dense (out.dense)). The table reports the performance for various fractions of retained singular vectors ($r_\text{frac}$) from the top, middle, and bottom subspaces.}
\label{tab:ablationdeltawfine-tune}
\end{table} 


\subsection{Fully Fine-tuned $\Delta W$}

In this experiment, $\Delta W$ represents the complete set of weight updates from fine-tuning. We fine-tuned RoBERTa-large on MRPC, SST-2, and MNLI, using a learning rate of 2e-5 for 5 epochs on MRPC and 3 epochs on SST-2 and MNLI. This experiment aims to assess the impact of retaining different fractions of the singular spectrum on performance for various weight modules, including query, key, value, intermediate dense, and output dense. The results are summarized in~\Cref{tab:ablationdeltawfine-tune}.

Our findings reveal that different modules exhibit varying sensitivity to the retention of singular vectors, suggesting that task-specific fine-tuning impacts each module differently. In the self-attention modules (query, key, value, and attention output), retaining even a small fraction of singular vectors (1\% or 10\%) resulted in minimal performance degradation across all tasks (in~\Cref{tab:ablationdeltawfine-tune} we only show results for the query because for other matrices the behavior was very similar). Both the top and bottom singular vectors maintained performance well. This may indicate that $\Delta W$ is relatively small for self-attention layers compared to $W$, meaning it has a limited influence on the model's predictions. This may suggest that these modules can tolerate substantial dimensionality reduction. The intermediate dense modules exhibited slightly higher sensitivity to the subspaces retained, though the impact remained small. 

The output dense modules demonstrated the greatest sensitivity to SVD-based approximations. This increased sensitivity in the output dense layers may indicate that $\Delta W/W$ may be larger in these layers. Consequently, we hypothesize that higher ranks should be used for the output dense layers, while lower ranks could be sufficient for the self-attention layers.

\clearpage

\section{Ablation Study: LoRA-XS initialization}
\label{appendix:ablation_loraxs_init}

In this ablation study, we examine the impact of different initialization strategies for the LoRA-XS matrices $A$ and $B$, comparing Singular Value Decomposition (SVD) initialization to random initialization (used in LoRA and VeRA). Additionally, we investigate whether applying SVD to random matrices yields any advantages, testing if the benefits of SVD initialization derive solely from the orthogonality of singular vectors, regardless of the source of the decomposition. 

For the SVD initialization, we decompose the original weight matrix as $W \approx U_r \Sigma_r V_r^T$, retaining the top $r$ singular vectors. Random initialization follows the Kaiming initialization, commonly used for linear layers in PyTorch~\citep{paszke2019pytorch}. 

We conduct our experiments using the RoBERTa-large~\citep{liu2019roberta} model on selected GLUE tasks. The LoRA-XS matrices are applied to specific transformer weights matrices, including the query, value, attention output, and intermediate output layers. All experiments are repeated with five different random seeds, and we report the median performance.

\subsection{Comparing Different Initializations}

We compare three initialization strategies: random initialization, SVD on random matrices (SVD of random), and SVD on the corresponding layer’s weight matrix (SVD of W). The results, summarized in \Cref{tab:ablation_init_mrpc}, \Cref{tab:ablation_init_cola}, and \Cref{tab:ablation_init_qnli}, reveal that initializing the $A$ and $B$ matrices with SVD on the corresponding module's original weight matrix ($W$) generally provides the best performance. This supports our theoretical hypothesis that leveraging the structure of the weight matrix through SVD initialization is more effective than random initialization, as it retains task-relevant information from the pretrained model (see~\Cref{theory}).

Our results also confirm that the performance benefits do not stem solely from the orthogonality of the singular vectors, as SVD on random matrices did not consistently outperform random initialization. 

Notably, for the SST-2 task (\Cref{tab:ablation_init_sst2}), all initialization methods perform similarly, with SVD on random matrices yielding the highest score. This is likely due to the nature of SST-2 as a sentiment classification task, which is less closely aligned with language modeling compared to other tasks such as MRPC, COLA, and QNLI. This finding aligns with our theoretical analysis (\Cref{theory}), which suggests that SVD on the original weights is most beneficial when the fine-tuning task is similar to the pretraining objective.

\begin{table*}[b]
\centering
\small
\begin{tabular}{|l|l|l|l|l|l|}
\hline
Rank & Init Type & LR & CLS LR & Median Score & Std Dev \\
\hline
4 & Random & 0.0005 & 0.0005 & 94.38 & 0.37 \\
4 & Random & 0.0005 & 0.001 & 94.50 & 0.37 \\
4 & Random & 0.0005 & 0.005 & 94.61 & 0.45 \\
4 & Random & 0.001 & 0.0005 & 94.61 & 0.41 \\
4 & Random & 0.001 & 0.001 & 94.72 & 0.36 \\
4 & Random & 0.001 & 0.005 & 94.61 & 0.33 \\
4 & Random & 0.005 & 0.0005 & 94.38 & 0.34 \\
4 & Random & 0.005 & 0.001 & 94.61 & 0.53 \\
4 & Random & 0.005 & 0.005 & 94.15 & 0.44 \\
\hline
4 & SVD of random & 0.0005 & 0.0005 & 94.15 & 0.81 \\
4 & SVD of random & 0.0005 & 0.001 & 94.04 & 0.63 \\
4 & SVD of random & 0.0005 & 0.005 & 93.92 & 0.92 \\
4 & SVD of random & 0.001 & 0.0005 & 94.38 & 0.65 \\
4 & SVD of random & 0.001 & 0.001 & 94.27 & 0.66 \\
4 & SVD of random & 0.001 & 0.005 & 94.04 & 0.85 \\
4 & SVD of random & 0.005 & 0.0005 & 94.72 & 0.54 \\
4 & \textbf{SVD of random} & 0.005 & 0.001 & \textbf{94.84} & 0.48 \\
4 & SVD of random & 0.005 & 0.005 & 94.50 & 0.50 \\
\hline
4 & SVD of W & 0.0005 & 0.0005 & 94.50 & 0.22 \\
4 & SVD of W & 0.0005 & 0.001 & 94.15 & 0.41 \\
4 & SVD of W & 0.0005 & 0.005 & 94.38 & 0.21 \\
4 & SVD of W & 0.001 & 0.0005 & 94.72 & 0.19 \\
4 & SVD of W & 0.001 & 0.001 & 94.38 & 0.09 \\
4 & SVD of W & 0.001 & 0.005 & 94.50 & 0.39 \\
4 & SVD of W & 0.005 & 0.0005 & 50.92 & 20.09 \\
4 & SVD of W & 0.005 & 0.001 & 88.99 & 19.32 \\
4 & SVD of W & 0.005 & 0.005 & 91.40 & 15.97 \\
\hline
\end{tabular}
\caption{Median accuracy scores for SST-2 across different initialization methods (rank 4). Despite similar performance across initialization strategies, the best score is achieved using SVD applied to random matrices, likely due to the task's nature as a sentiment classification challenge.}
\label{tab:ablation_init_sst2}
\end{table*}

\begin{table*}[h]
\centering
\small
\begin{tabular}{|l|l|l|l|l|l|}
\hline
Rank & Init Type & LR & CLS LR & Median Score & Std Dev \\
\hline
4 & Random & 0.0005 & 0.0005 & 82.60 & 0.63 \\
4 & Random & 0.0005 & 0.001 & 82.84 & 0.93 \\
4 & Random & 0.0005 & 0.005 & 83.33 & 0.78 \\
4 & Random & 0.001 & 0.0005 & 84.80 & 0.63 \\
4 & Random & 0.001 & 0.001 & 84.80 & 1.04 \\
4 & Random & 0.001 & 0.005 & 84.80 & 1.25 \\
4 & Random & 0.005 & 0.0005 & 86.76 & 0.70 \\
4 & Random & 0.005 & 0.001 & 85.78 & 0.80 \\
4 & Random & 0.005 & 0.005 & 86.76 & 0.96 \\
\hline
4 & SVD of random & 0.0005 & 0.0005 & 78.92 & 1.25 \\
4 & SVD of random & 0.0005 & 0.001 & 79.41 & 1.36 \\
4 & SVD of random & 0.0005 & 0.005 & 78.43 & 1.05 \\
4 & SVD of random & 0.001 & 0.0005 & 81.62 & 0.77 \\
4 & SVD of random & 0.001 & 0.001 & 81.13 & 1.19 \\
4 & SVD of random & 0.001 & 0.005 & 80.64 & 1.61 \\
4 & SVD of random & 0.005 & 0.0005 & 84.31 & 0.74 \\
4 & SVD of random & 0.005 & 0.001 & 84.31 & 1.07 \\
4 & SVD of random & 0.005 & 0.005 & 84.56 & 1.27 \\
\hline
4 & SVD of W & 0.0005 & 0.0005 & 86.76 & 0.68 \\
4 & SVD of W & 0.0005 & 0.001 & 86.52 & 0.33 \\
4 & SVD of W & 0.0005 & 0.005 & 86.03 & 0.87 \\
4 & SVD of W & 0.001 & 0.0005 & 87.50 & 0.73 \\
4 & SVD of W & 0.001 & 0.001 & 87.25 & 0.72 \\
4 & \textbf{SVD of W} & 0.001 & 0.005 & \textbf{87.50} & 0.99 \\
4 & SVD of W & 0.005 & 0.0005 & 69.12 & 2.13 \\
4 & SVD of W & 0.005 & 0.001 & 70.59 & 2.05 \\
4 & SVD of W & 0.005 & 0.005 & 68.38 & 2.27 \\
\hline
\end{tabular}
\caption{Median accuracy scores for MRPC across different initialization methods (rank 4). SVD applied to the original weight matrix provides the best performance, aligning with the hypothesis that initializing based on weight-specific information better adapts LoRA-XS to the task.}
\label{tab:ablation_init_mrpc}
\end{table*}

\begin{table*}[h]
\centering
\small
\begin{tabular}{|l|l|l|l|l|l|}
\hline
Rank & Init Type & LR & CLS LR & Median Score & Std Dev \\
\hline
4 & Random & 0.0005 & 0.0005 & 51.88 & 0.83 \\
4 & Random & 0.0005 & 0.001 & 53.12 & 1.05 \\
4 & Random & 0.0005 & 0.005 & 54.81 & 2.08 \\
4 & Random & 0.001 & 0.0005 & 52.90 & 1.63 \\
4 & Random & 0.001 & 0.001 & 53.38 & 1.12 \\
4 & Random & 0.001 & 0.005 & 58.53 & 1.84 \\
4 & Random & 0.005 & 0.0005 & 52.48 & 1.44 \\
4 & Random & 0.005 & 0.001 & 53.17 & 1.88 \\
4 & Random & 0.005 & 0.005 & 56.34 & 1.15 \\
\hline
4 & SVD of random & 0.0005 & 0.0005 & 47.96 & 1.52 \\
4 & SVD of random & 0.0005 & 0.001 & 48.58 & 2.02 \\
4 & SVD of random & 0.0005 & 0.005 & 53.30 & 0.92 \\
4 & SVD of random & 0.001 & 0.0005 & 50.14 & 1.88 \\
4 & SVD of random & 0.001 & 0.001 & 51.47 & 2.06 \\
4 & SVD of random & 0.001 & 0.005 & 55.27 & 0.96 \\
4 & SVD of random & 0.005 & 0.0005 & 52.33 & 2.72 \\
4 & SVD of random & 0.005 & 0.001 & 53.32 & 2.16 \\
4 & SVD of random & 0.005 & 0.005 & 56.33 & 1.21 \\
\hline
4 & SVD of W & 0.0005 & 0.0005 & 55.99 & 0.55 \\
4 & SVD of W & 0.0005 & 0.001 & 55.74 & 1.17 \\
4 & SVD of W & 0.0005 & 0.005 & 58.20 & 1.66 \\
4 & SVD of W & 0.001 & 0.0005 & 57.63 & 0.78 \\
4 & SVD of W & 0.001 & 0.001 & 56.76 & 0.66 \\
4 & \textbf{SVD of W} & 0.001 & 0.005 & \textbf{60.11} & 0.85 \\
4 & SVD of W & 0.005 & 0.0005 & 43.10 & 7.78 \\
4 & SVD of W & 0.005 & 0.001 & 38.02 & 16.30 \\
4 & SVD of W & 0.005 & 0.005 & 45.87 & 8.41 \\
\hline
\end{tabular}
\caption{Median performance scores (Matthews correlation) for COLA across different initialization methods (rank 4). SVD applied to the original weight matrix provides the best performance, aligning with the hypothesis that initializing based on weight-specific information better adapts LoRA-XS to the task.}
\label{tab:ablation_init_cola}
\end{table*}

\begin{table*}[h]
\centering
\small
\begin{tabular}{|l|l|l|l|l|l|}
\hline
Rank & Init Type & LR & CLS LR & Median Score & Std Dev \\
\hline
4 & Random & 0.0005 & 0.0005 & 87.92 & 0.56 \\
4 & Random & 0.0005 & 0.001 & 87.64 & 0.31 \\
4 & Random & 0.0005 & 0.005 & 87.79 & 0.32 \\
4 & Random & 0.001 & 0.0005 & 88.17 & 0.44 \\
4 & Random & 0.001 & 0.001 & 88.45 & 0.27 \\
4 & Random & 0.001 & 0.005 & 87.96 & 0.46 \\
4 & Random & 0.005 & 0.0005 & 88.91 & 0.56 \\
4 & Random & 0.005 & 0.001 & 88.80 & 0.42 \\
4 & Random & 0.005 & 0.005 & 88.32 & 0.35 \\
\hline
4 & SVD of random & 0.0005 & 0.0005 & 84.48 & 1.14 \\
4 & SVD of random & 0.0005 & 0.001 & 84.42 & 1.00 \\
4 & SVD of random & 0.0005 & 0.005 & 83.12 & 0.98 \\
4 & SVD of random & 0.001 & 0.0005 & 86.36 & 0.88 \\
4 & SVD of random & 0.001 & 0.001 & 86.18 & 1.01 \\
4 & SVD of random & 0.001 & 0.005 & 85.30 & 0.97 \\
4 & SVD of random & 0.005 & 0.0005 & 88.36 & 0.58 \\
4 & SVD of random & 0.005 & 0.001 & 88.34 & 0.42 \\
4 & SVD of random & 0.005 & 0.005 & 87.33 & 0.64 \\
\hline
4 & SVD of W & 0.0005 & 0.0005 & 90.74 & 0.19 \\
4 & SVD of W & 0.0005 & 0.001 & 90.74 & 0.13 \\
4 & SVD of W & 0.0005 & 0.005 & 90.41 & 0.10 \\
4 & \textbf{SVD of W} & 0.001 & 0.0005 & \textbf{90.94} & 0.20 \\
4 & SVD of W & 0.001 & 0.001 & 90.92 & 0.17 \\
4 & SVD of W & 0.001 & 0.005 & 90.59 & 0.90 \\
4 & SVD of W & 0.005 & 0.0005 & 50.54 & 14.37 \\
4 & SVD of W & 0.005 & 0.001 & 50.54 & 0.00 \\
4 & SVD of W & 0.005 & 0.005 & 50.54 & 0.00 \\
\hline
\end{tabular}
\caption{Median accuracy scores for QNLI across different initialization methods (rank 4). SVD applied to the original weight matrix provides the best performance, aligning with the hypothesis that initializing based on weight-specific information better adapts LoRA-XS to the task.}
\label{tab:ablation_init_qnli}
\end{table*}

\clearpage

\subsection{Effect of Initialization on Learning Curve and Rank}

We further explore the impact of initialization on learning speed and performance at different ranks.

In~\Cref{tab:ablationsvdvsnosvd}, we present detailed results comparing the initialization of the $A$ and $B$ matrices of LoRA-XS using SVD versus random initialization across various ranks. We also report results after 1 and 2 epochs to illustrate the learning curve and how quickly the models learn with each initialization method. The reported metrics are accuracy for SST-2 and QNLI and Matthews correlation for CoLA on the RoBERTa-large model. Hyperparameters were selected through grid search and are detailed in~\Cref{tab:gluesvdnosvdhyperparams}. Results are reported as the median over 5 random seeds, with the best epoch result for each run shown in the "Best epoch" column. 

Our results empirically confirm that, across different ranks, aligning adaptation matrices with the principal components of pre-trained weights is beneficial compared to random initialization. As shown in~\Cref{tab:ablationsvdvsnosvd}, initializing the $A$ and $B$ matrices with SVD generally leads to improved performance.

Moreover, the performance after 1 and 2 epochs is often higher with SVD initialization. This suggests that proper initialization using SVD may help the models converge more quickly and achieve better overall accuracy. This finding underscores the importance of initialization strategy in fine-tuning large language models, highlighting that SVD initialization not only boosts performance but also accelerates the training process.

\begin{table*}[t!]
    \small
    \centering
    \vskip 0.1in
    \begin{tabular}{@{}cccccc@{}}

\toprule
Task & Rank & Init Type & \multicolumn{3}{c}{Performance} \\
\cmidrule(lr){4-6}
 & & & After 1 epoch & After 2 epoch & Best epoch \\
\midrule

\multirow{8}{*}{Cola} 

 & \multirow{2}{*}{8} & Random & 0 $\pm$ 17.59 & 40.91 $\pm$ 15.57 & 60.29 $\pm$ 1.22  \\
 & & SVD of W & \textbf{12.59} $\pm$ 11.09 & \textbf{43.84} $\pm$ 15.62 & \textbf{64.39} $\pm$ 0.75 \\ \cmidrule(lr){2-6}
 & \multirow{2}{*}{12} & Random & 5.8 $\pm$ 14.71 & 26.77 $\pm$ 20.43 & 62.96 $\pm$ 1.25  \\
 & & SVD of W & \textbf{13.16} $\pm$ 17.84 & \textbf{44.22} $\pm$ 13.04 & \textbf{65.47} $\pm$ 0.90 \\ \cmidrule(lr){2-6}
 & \multirow{2}{*}{20} & Random & \textbf{47.17} $\pm$ 2.59 & 43.42 $\pm$ 5.70 & 64.7 $\pm$ 1.10  \\
 & & SVD of W & 18.83 $\pm$ 9.47 & \textbf{44.77} $\pm$ 9.65 & \textbf{68.08} $\pm$ 1.21 \\

\midrule

\multirow{8}{*}{SST-2} 

 & \multirow{2}{*}{8} 
 & Random & 92.55 $\pm$ 0.74 & 93.58 $\pm$ 0.53 & \textbf{95.30} $\pm$ 0.34 \\
 & & SVD of W & \textbf{93.23} $\pm$ 0.27 & \textbf{93.81} $\pm$ 0.22 & 94.95 $\pm$ 0.07 \\ \cmidrule(lr){2-6}
 & \multirow{2}{*}{12} & Random & \textbf{93.46} $\pm$ 0.54 & \textbf{94.27} $\pm$ 0.46 & 95.64 $\pm$ 0.54  \\
 & & SVD of W & \textbf{93.46} $\pm$ 0.86 & 93.35 $\pm$ 0.49 & \textbf{95.87} $\pm$ 0.28  \\ \cmidrule(lr){2-6}
 & \multirow{2}{*}{20} & Random & 94.15 $\pm$ 0.67 & \textbf{94.15} $\pm$ 0.58 & \textbf{95.87} $\pm$ 0.24  \\
 & & SVD of W & \textbf{94.61} $\pm$ 0.46 & 93.92 $\pm$ 0.87 & \textbf{95.87} $\pm$ 0.31  \\

\midrule

\multirow{8}{*}{QNLI} 

 & \multirow{2}{*}{8} & Random & 88.5 $\pm$ 0.86 & 89.99 $\pm$ 0.72 & 91.4 $\pm$ 0.59 \\
 & & SVD of W & \textbf{89.84} $\pm$ 0.37 & \textbf{91.31} $\pm$ 0.10 & \textbf{92.49} $\pm$ 0.09  \\ \cmidrule(lr){2-6}
 & \multirow{2}{*}{12} & Random & 89.27 $\pm$ 0.55 & 90.3 $\pm$ 0.58 & 92.51 $\pm$ 0.42 \\
 & & SVD of W & \textbf{90.74} $\pm$ 0.25 & \textbf{92.17} $\pm$ 0.53 & \textbf{93.32} $\pm$ 0.51  \\ \cmidrule(lr){2-6}
 & \multirow{2}{*}{20} & Random & 89.73 $\pm$ 0.96 & 91.45 $\pm$ 0.11 & 93.23 $\pm$ 0.30 \\
 & & SVD of W & \textbf{91.49} $\pm$ 0.18 & \textbf{92.24} $\pm$ 0.24 & \textbf{94.05} $\pm$ 0.16  \\
 \bottomrule

    \end{tabular}
    \caption{The impact of random vs SVD of W initialization scheme on the performance of RoBERTa-large on three GLUE tasks across different ranks. Matrices $A$ and $B$ in LoRA-XS are initialized either randomly or using SVD of the corresponding weight.  We report Matthew's Correlation for CoLA and accuracy for SST-2 and QNLI tasks. Each table entry reports median and standard deviation of five runs with different seeds. Our study indicates that the SVD-based initialization generally results in better performance, especially when dealing with lower ranks (\ie, less trainable parameters).}
    \label{tab:ablationsvdvsnosvd}
\end{table*}

\begin{table*}[h]
    \small
    \centering
    \vskip 0.1in
    \begin{tabular}{@{}c|c|c|c|c|c@{}}
         \midrule
         Task & Init & Rank & LoRA-XS LR & Classifier LR & Epochs \\
         \midrule
         \multirow{6}{*}{SST-2} 
         & \multirow{3}{*}{SVD of W} 
         & 8 & 1E-3 & 1E-3 & 20 \\
         &                     & 12 & 5E-3 & 1E-3 & 20 \\
         &                     & 20 & 1E-3 & 5E-3 & 20 \\
         \cline{2-6}
         & \multirow{3}{*}{Random} 
         & 8 & 1E-3 & 1E-3 & 20 \\
         &                          & 12 & 1E-3 & 1E-3 & 20 \\
         &                          & 20 & 5E-3 & 1E-4 & 20 \\
         \midrule

         \multirow{6}{*}{CoLA} 
         & \multirow{3}{*}{SVD of W} 
         & 8 & 1E-3 & 5E-3 & 50 \\
         &                     & 12 & 1E-3 & 5E-3 & 50 \\
         &                     & 20 & 1E-3 & 5E-3 & 50 \\
         \cline{2-6}
         & \multirow{3}{*}{Random} 
         & 8 & 1E-3 & 1E-2 & 50 \\
         &                          & 12 & 1E-3 & 1E-2 & 50 \\
         &                          & 20 & 5E-3 & 5E-3 & 50 \\
         \midrule

         \multirow{6}{*}{QNLI} 
         & \multirow{3}{*}{SVD of W} 
        & 8 & 1E-3 & 1E-3 & 10 \\
         &                     & 12 & 1E-3 & 5E-4 & 10 \\
         &                     & 20 & 1E-3 & 5E-4 & 10 \\
         \cline{2-6}
         & \multirow{3}{*}{Random} 
         & 8 & 1E-3 & 1E-3 & 10 \\
         &                          & 12 & 1E-3 & 5E-4 & 10 \\
         &                          & 20 & 1E-3 & 1E-3 & 10 \\
         
         \bottomrule
    \end{tabular}
    \vspace{10pt}
    \caption{Hyperparameters for the ablation study with different $A$ and $B$ initialization methods (SVD and random) on RoBERTa-large with LoRA-XS across various GLUE tasks.}
    \label{tab:gluesvdnosvdhyperparams}
\end{table*}

\clearpage

\section{Ablation Study: Top vs. Bottom Singular Vector Initialization}
\label{ablation:topvsbottomsingularvectorsinit}

In this ablation study, we evaluate the effect of initializing LoRA-XS with either top or bottom singular vectors derived from the SVD of the weight matrix $W$. We conduct experiments on the RoBERTa-large model across multiple GLUE benchmark tasks, including CoLA, QNLI, MRPC, and SST-2. The target layers for LoRA-XS in these experiments are the query, value, attention.output.dense, and output.dense layers. We set the LoRA scaling factor $\alpha$ to 16 and the rank $r$ to 4.

Given a weight matrix $W \in \mathbb{R}^{m \times n}$, we compute its SVD as $W = U \Sigma V^T$, where $U \in \mathbb{R}^{m \times m}$, $\Sigma \in \mathbb{R}^{m \times n}$ is a diagonal matrix of singular values, and $V \in \mathbb{R}^{n \times n}$.

For LoRA-XS initialization, we modify the decomposition by selecting either the top or bottom $r$ singular vectors from $U$ and $V$:

\begin{itemize}
    \item \textbf{Top SVD initialization:} We select the first $r$ singular vectors from $U$ and $V$. Specifically, $A = U_r \Sigma_r$ and $B = V_r^T$, where $U_r \in \mathbb{R}^{m \times r}$ and $V_r \in \mathbb{R}^{n \times r}$ represent the top $r$ singular vectors.
    \item \textbf{Bottom SVD initialization:} We select the last $r$ singular vectors from $U$ and $V$, i.e., $A = U_{-r} \Sigma_{-r}$ and $B = V_{-r}^T$.
\end{itemize}

Each configuration is tested over five random seeds, and we report the median performance. The results are summarized in \Cref{tab:topvsbottom_cola}, \Cref{tab:topvsbottom_qnli}, \Cref{tab:topvsbottom_mrpc}, and \Cref{tab:topvsbottom_sst2}.

Across all tasks, initializing LoRA-XS with top singular vectors consistently outperforms initialization with bottom singular vectors. These results support our decision to adopt top singular vector initialization for LoRA-XS.

\begin{table*}[h!]
\centering
\small
\begin{tabular}{|c|c|c|c|c|c|}
\hline
Rank & Init Type & LR & CLS LR & Median Score & Std Dev \\
\hline
4 & svd on w bottom & 0.0005 & 0.0005 & 42.26 & 1.71 \\
4 & svd on w bottom & 0.0005 & 0.001 & 43.28 & 1.46 \\
4 & svd on w bottom & 0.0005 & 0.005 & 46.09 & 1.26 \\
4 & svd on w bottom & 0.001 & 0.0005 & 45.24 & 1.71 \\
4 & svd on w bottom & 0.001 & 0.001 & 44.53 & 0.60 \\
4 & svd on w bottom & 0.001 & 0.005 & 48.24 & 0.62 \\
4 & svd on w bottom & 0.005 & 0.0005 & 47.04 & 1.06 \\
4 & svd on w bottom & 0.005 & 0.001 & 48.89 & 0.83 \\
4 & svd on w bottom & 0.005 & 0.005 & 52.01 & 0.92 \\
4 & svd on w top & 0.0005 & 0.0005 & 55.01 & 0.58 \\
4 & svd on w top & 0.0005 & 0.001 & 55.47 & 0.63 \\
4 & svd on w top & 0.0005 & 0.005 & 58.63 & 1.21 \\
4 & svd on w top & 0.001 & 0.0005 & 58.04 & 1.31 \\
4 & svd on w top & 0.001 & 0.001 & 57.46 & 0.75 \\
4 & \textbf{svd on w top} & 0.001 & 0.005 & \textbf{60.29} & 1.06 \\
4 & svd on w top & 0.005 & 0.0005 & 43.78 & 14.36 \\
4 & svd on w top & 0.005 & 0.001 & 47.60 & 3.72 \\
4 & svd on w top & 0.005 & 0.005 & 25.51 & 15.58 \\
\hline
\end{tabular}
\caption{Results for CoLA task comparing LoRA-XS initialized with top versus bottom singular vectors across the query, value, attention.output.dense, and "output.dense" modules in RoBERTa-large model. Top singular vectors demonstrate superior performance.}
\label{tab:topvsbottom_cola}
\end{table*}

\begin{table*}[h!]
\centering
\small
\begin{tabular}{|l|l|l|l|l|l|}
\hline
Rank & Init Type & LR & CLS LR & Median Score & Std Dev \\
\hline
4 & svd on w bottom & 0.0005 & 0.0005 & 73.88 & 0.27 \\
4 & svd on w bottom & 0.0005 & 0.001 & 73.24 & 0.23 \\
4 & svd on w bottom & 0.0005 & 0.005 & 66.17 & 1.67 \\
4 & svd on w bottom & 0.001 & 0.0005 & 77.52 & 0.25 \\
4 & svd on w bottom & 0.001 & 0.001 & 77.14 & 0.33 \\
4 & svd on w bottom & 0.001 & 0.005 & 70.14 & 1.13 \\
4 & svd on w bottom & 0.005 & 0.0005 & 81.29 & 0.29 \\
4 & svd on w bottom & 0.005 & 0.001 & 81.38 & 0.22 \\
4 & svd on w bottom & 0.005 & 0.005 & 79.90 & 0.33 \\
4 & svd on w top & 0.0005 & 0.0005 & 90.88 & 0.15 \\
4 & svd on w top & 0.0005 & 0.001 & 90.68 & 0.05 \\
4 & svd on w top & 0.0005 & 0.005 & 90.28 & 0.20 \\
4 & \textbf{svd on w top} & 0.001 & 0.0005 & \textbf{90.98} & 0.09 \\
4 & svd on w top & 0.001 & 0.001 & 90.96 & 0.11 \\
4 & svd on w top & 0.001 & 0.005 & 90.72 & 0.21 \\
4 & svd on w top & 0.005 & 0.0005 & 50.54 & 0.75 \\
4 & svd on w top & 0.005 & 0.001 & 50.54 & 0.00 \\
4 & svd on w top & 0.005 & 0.005 & 50.54 & 0.00 \\
\hline
\end{tabular}
\caption{Results for QNLI task comparing LoRA-XS initialized with top versus bottom singular vectors across the query, value, attention.output.dense, and "output.dense" modules in RoBERTa-large model. Top singular vectors demonstrate superior performance.}
\label{tab:topvsbottom_qnli}
\end{table*}

\begin{table*}[h!]
\centering
\small
\begin{tabular}{|c|c|c|c|c|c|}
\hline
Rank & Init Type & LR & CLS LR & Median Score & Std Dev \\
\hline
4 & svd on w bottom & 0.0005 & 0.0005 & 75.25 & 0.66 \\
4 & svd on w bottom & 0.0005 & 0.001 & 75.49 & 0.42 \\
4 & svd on w bottom & 0.0005 & 0.005 & 74.51 & 0.48 \\
4 & svd on w bottom & 0.001 & 0.0005 & 75.00 & 0.50 \\
4 & svd on w bottom & 0.001 & 0.001 & 75.49 & 0.47 \\
4 & svd on w bottom & 0.001 & 0.005 & 75.00 & 0.29 \\
4 & svd on w bottom & 0.005 & 0.0005 & 78.92 & 0.48 \\
4 & svd on w bottom & 0.005 & 0.001 & 79.17 & 0.48 \\
4 & svd on w bottom & 0.005 & 0.005 & 77.94 & 0.41 \\
4 & svd on w top & 0.0005 & 0.0005 & 86.27 & 0.63 \\
4 & svd on w top & 0.0005 & 0.001 & 85.54 & 1.01 \\
4 & svd on w top & 0.0005 & 0.005 & 85.54 & 1.23 \\
4 & svd on w top & 0.001 & 0.0005 & 86.52 & 0.73 \\
4 & \textbf{svd on w top} & 0.001 & 0.001 & \textbf{86.76} & 1.11 \\
4 & \textbf{svd on w top} & 0.001 & 0.005 & \textbf{86.76} & 0.81 \\
4 & svd on w top & 0.005 & 0.0005 & 69.85 & 6.25 \\
4 & svd on w top & 0.005 & 0.001 & 74.26 & 6.69 \\
4 & svd on w top & 0.005 & 0.005 & 68.63 & 1.79 \\
\hline
\end{tabular}
\caption{Results for MRPC task comparing LoRA-XS initialized with top versus bottom singular vectors across the query, value, attention.output.dense, and "output.dense" modules in RoBERTa-large model. Top singular vectors demonstrate superior performance.}
\label{tab:topvsbottom_mrpc}
\end{table*}

\begin{table*}[h!]
\centering
\small
\begin{tabular}{|c|c|c|c|c|c|}
\hline
Rank & Init Type & LR & CLS LR & Median Score & Std Dev \\
\hline
4 & svd on w bottom & 0.0005 & 0.0005 & 89.11 & 0.16 \\
4 & svd on w bottom & 0.0005 & 0.001 & 89.45 & 0.28 \\
4 & svd on w bottom & 0.0005 & 0.005 & 88.76 & 0.09 \\
4 & svd on w bottom & 0.001 & 0.0005 & 90.48 & 0.11 \\
4 & svd on w bottom & 0.001 & 0.001 & 90.94 & 0.22 \\
4 & svd on w bottom & 0.001 & 0.005 & 90.08 & 2.04 \\
4 & svd on w bottom & 0.005 & 0.0005 & 92.89 & 0.28 \\
4 & svd on w bottom & 0.005 & 0.001 & 93.12 & 0.12 \\
4 & svd on w bottom & 0.005 & 0.005 & 92.66 & 0.12 \\
4 & \textbf{svd on w top} & 0.0005 & 0.0005 & \textbf{94.61} & 0.17 \\
4 & svd on w top & 0.0005 & 0.001 & 94.38 & 0.15 \\
4 & svd on w top & 0.0005 & 0.005 & 94.50 & 0.15 \\
4 & svd on w top & 0.001 & 0.0005 & 94.50 & 0.14 \\
4 & svd on w top & 0.001 & 0.001 & 94.38 & 0.16 \\
4 & svd on w top & 0.001 & 0.005 & 94.21 & 0.10 \\
4 & svd on w top & 0.005 & 0.0005 & 91.86 & 20.21 \\
4 & svd on w top & 0.005 & 0.001 & 91.28 & 16.27 \\
4 & svd on w top & 0.005 & 0.005 & 50.92 & 19.47 \\
\hline
\end{tabular}
\caption{Results for SST-2 task comparing LoRA-XS initialized with top versus bottom singular vectors across the query, value, attention.output.dense, and "output.dense" modules in RoBERTa-large model. Top singular vectors demonstrate superior performance.}
\label{tab:topvsbottom_sst2}
\end{table*}

\clearpage

\section{Ablation Study: The Importance of Singular Values for LoRA-XS}
\label{ablation:sigmaornotinit}

In this section, we explore the significance of including singular values in the initialization of matrix $A$ in LoRA-XS. Specifically, we compare two initialization methods: one where matrix $A$ is initialized with both singular vectors and singular values, $A = U \Sigma$, and another where matrix $A$ is initialized using only the singular vectors, $A = U$, while keeping $B = V^T$ in both cases. Our analysis focuses on the top $r$ singular vectors for both methods.

For LoRA-XS, we initialize the matrices $A$ and $B$ as:

\begin{equation}
A = U_r \Sigma_r \quad \text{and} \quad B = V_r^T.
\end{equation}

In the variant without singular values, matrix $A$ is initialized as:

\begin{equation}
A = U_r \quad \text{and} \quad B = V_r^T.
\end{equation}

We conduct experiments using the RoBERTa-large model across several GLUE benchmark tasks, including CoLA, QNLI, MRPC, and SST-2. The target layers for LoRA-XS are the query, value, attention.output.dense, and output.dense layers. We set the LoRA scaling factor $\alpha$ to 16 and use a rank of 4. Each configuration is tested over 5 random seeds, and we report the median performance. The results are presented in \Cref{tab:sst2_sigma}, \Cref{tab:mrpc_sigma}, \Cref{tab:cola_sigma}, and \Cref{tab:qnli_sigma}.

Overall, our results indicate that including singular values ($A = U_r\Sigma_r$) provides consistent benefits across most tasks, particularly for CoLA, QNLI, and SST-2. However, for the MRPC task, initializing matrix $A$ without the singular values ($A = U_r$) achieves better performance. The findings support the inclusion of singular values in most cases, but also demonstrate that certain tasks, like MRPC, may benefit from simpler initialization methods.

\begin{table*}[h!]
\centering
\small
\begin{tabular}{|l|l|l|l|l|l|}
\hline
Rank & Init Type & LR & CLS LR & Median Score & Std Dev \\
\hline
4 & $A = U_r\Sigma_r$ & 0.0005 & 0.0005 & \textbf{94.61} & 0.17 \\
4 & $A = U_r\Sigma_r$ & 0.0005 & 0.001 & 94.38 & 0.15 \\
4 & $A = U_r\Sigma_r$ & 0.0005 & 0.005 & 94.50 & 0.15 \\
4 & $A = U_r\Sigma_r$ & 0.001 & 0.0005 & 94.50 & 0.14 \\
4 & $A = U_r\Sigma_r$ & 0.001 & 0.001 & 94.38 & 0.16 \\
4 & $A = U_r\Sigma_r$ & 0.001 & 0.005 & 94.21 & 0.10 \\
4 & $A = U_r\Sigma_r$ & 0.005 & 0.0005 & 91.86 & 20.21 \\
4 & $A = U_r\Sigma_r$ & 0.005 & 0.001 & 91.28 & 16.27 \\
4 & $A = U_r\Sigma_r$ & 0.005 & 0.005 & 50.92 & 19.47 \\
\hline
4 & $A = U_r$ & 0.0005 & 0.0005 & 93.81 & 0.05 \\
4 & $A = U_r$ & 0.0005 & 0.001 & 93.69 & 0.16 \\
4 & $A = U_r$ & 0.0005 & 0.005 & 93.92 & 0.14 \\
4 & $A = U_r$ & 0.001 & 0.0005 & 93.69 & 0.20 \\
4 & $A = U_r$ & 0.001 & 0.001 & 94.04 & 0.16 \\
4 & $A = U_r$ & 0.001 & 0.005 & 93.92 & 0.19 \\
4 & $A = U_r$ & 0.005 & 0.0005 & 94.38 & 0.32 \\
4 & $A = U_r$ & 0.005 & 0.001 & 94.15 & 0.34 \\
4 & $A = U_r$ & 0.005 & 0.005 & 94.27 & 0.24 \\
\hline
\end{tabular}
\caption{Comparison of LoRA-XS initialization with and without singular values on SST-2.}
\label{tab:sst2_sigma}
\end{table*}

\begin{table*}[h!]
\centering
\small
\begin{tabular}{|l|l|l|l|l|l|}
\hline
Rank & Init Type & LR & CLS LR & Median Score & Std Dev \\
\hline
4 & $A = U_r\Sigma_r$ & 0.0005 & 0.0005 & 86.27 & 0.63 \\
4 & $A = U_r\Sigma_r$ & 0.0005 & 0.001 & 85.54 & 1.01 \\
4 & $A = U_r\Sigma_r$ & 0.0005 & 0.005 & 85.54 & 1.23 \\
4 & $A = U_r\Sigma_r$ & 0.001 & 0.0005 & 86.52 & 0.73 \\
4 & $A = U_r\Sigma_r$ & 0.001 & 0.001 & 86.76 & 1.11 \\
4 & $A = U_r\Sigma_r$ & 0.001 & 0.005 & 86.76 & 0.81 \\
4 & $A = U_r\Sigma_r$ & 0.005 & 0.0005 & 69.85 & 6.25 \\
4 & $A = U_r\Sigma_r$ & 0.005 & 0.001 & 74.26 & 6.69 \\
4 & $A = U_r\Sigma_r$ & 0.005 & 0.005 & 68.63 & 1.79 \\
\hline
4 & $A = U_r$ & 0.0005 & 0.0005 & 82.60 & 0.75 \\
4 & $A = U_r$ & 0.0005 & 0.001 & 83.33 & 0.67 \\
4 & $A = U_r$ & 0.0005 & 0.005 & 82.11 & 0.50 \\
4 & $A = U_r$ & 0.001 & 0.0005 & 86.03 & 0.57 \\
4 & $A = U_r$ & 0.001 & 0.001 & 86.52 & 0.77 \\
4 & $A = U_r$ & 0.001 & 0.005 & 85.78 & 0.25 \\
4 & $A = U_r$ & 0.005 & 0.0005 & 87.50 & 0.50 \\
4 & $A = U_r$ & 0.005 & 0.001 & \textbf{88.24} & 0.24 \\
4 & $A = U_r$ & 0.005 & 0.005 & 87.50 & 1.27 \\
\hline
\end{tabular}
\caption{Comparison of LoRA-XS initialization with and without singular values on MRPC.}
\label{tab:mrpc_sigma}
\end{table*}

\begin{table*}[h!]
\centering
\small
\begin{tabular}{|l|l|l|l|l|l|}
\hline
Rank & Init Type & LR & CLS LR & Median Score & Std Dev \\
\hline
4 & $A = U_r\Sigma_r$ & 0.0005 & 0.0005 & 55.01 & 0.58 \\
4 & $A = U_r\Sigma_r$ & 0.0005 & 0.001 & 55.47 & 0.63 \\
4 & $A = U_r\Sigma_r$ & 0.0005 & 0.005 & 58.63 & 1.21 \\
4 & $A = U_r\Sigma_r$ & 0.001 & 0.0005 & 58.04 & 1.31 \\
4 & $A = U_r\Sigma_r$ & 0.001 & 0.001 & 57.46 & 0.75 \\
4 & \textbf{$A = U_r\Sigma_r$} & 0.001 & 0.005 & \textbf{60.29} & 1.06 \\
4 & $A = U_r\Sigma_r$ & 0.005 & 0.0005 & 43.78 & 14.36 \\
4 & $A = U_r\Sigma_r$ & 0.005 & 0.001 & 47.60 & 3.72 \\
4 & $A = U_r\Sigma_r$ & 0.005 & 0.005 & 25.51 & 15.58 \\
\hline
4 & $A = U_r$ & 0.0005 & 0.0005 & 51.21 & 0.67 \\
4 & $A = U_r$ & 0.0005 & 0.001 & 52.07 & 0.90 \\
4 & $A = U_r$ & 0.0005 & 0.005 & 54.06 & 0.63 \\
4 & $A = U_r$ & 0.001 & 0.0005 & 52.59 & 0.33 \\
4 & $A = U_r$ & 0.001 & 0.001 & 52.40 & 0.65 \\
4 & $A = U_r$ & 0.001 & 0.005 & 55.52 & 0.73 \\
4 & $A = U_r$ & 0.005 & 0.0005 & 56.79 & 1.08 \\
4 & $A = U_r$ & 0.005 & 0.001 & 57.22 & 0.52 \\
4 & $A = U_r$ & 0.005 & 0.005 & 58.74 & 0.63 \\
\hline
\end{tabular}
\caption{Comparison of LoRA-XS initialization with and without singular values on CoLA.}
\label{tab:cola_sigma}
\end{table*}

\begin{table*}[h!]
\centering
\small
\begin{tabular}{|l|l|l|l|l|l|}
\hline
Rank & Init Type & LR & CLS LR & Median Score & Std Dev \\
\hline
4 & $A = U_r\Sigma_r$ & 0.0005 & 0.0005 & 90.88 & 0.15 \\
4 & $A = U_r\Sigma_r$ & 0.0005 & 0.001 & 90.68 & 0.05 \\
4 & $A = U_r\Sigma_r$ & 0.0005 & 0.005 & 90.28 & 0.20 \\
4 & \textbf{$A = U_r\Sigma_r$} & 0.001 & 0.0005 & \textbf{90.98} & 0.09 \\
4 & $A = U_r\Sigma_r$ & 0.001 & 0.001 & 90.96 & 0.11 \\
4 & $A = U_r\Sigma_r$ & 0.001 & 0.005 & 90.72 & 0.21 \\
4 & $A = U_r\Sigma_r$ & 0.005 & 0.0005 & 50.54 & 0.75 \\
4 & $A = U_r\Sigma_r$ & 0.005 & 0.001 & 50.54 & 0.00 \\
4 & $A = U_r\Sigma_r$ & 0.005 & 0.005 & 50.54 & 0.00 \\
\hline
4 & $A = U_r$ & 0.0005 & 0.0005 & 88.43 & 0.19 \\
4 & $A = U_r$ & 0.0005 & 0.001 & 85.30 & 0.00 \\
4 & $A = U_r$ & 0.001 & 0.0005 & 89.86 & 0.18 \\
4 & $A = U_r$ & 0.001 & 0.001 & 89.80 & 0.21 \\
4 & $A = U_r$ & 0.001 & 0.005 & 89.07 & 0.50 \\
4 & $A = U_r$ & 0.005 & 0.0005 & 90.87 & 0.01 \\
4 & $A = U_r$ & 0.005 & 0.001 & 90.92 & 0.15 \\
4 & $A = U_r$ & 0.005 & 0.005 & 90.48 & 0.29 \\
\hline
\end{tabular}
\caption{Comparison of LoRA-XS initialization with and without singular values on QNLI.}
\label{tab:qnli_sigma}
\end{table*}

\end{document}